\documentclass[10pt,twocolumn,letterpaper,dvipsnames]{article}

\usepackage[pagenumbers]{wacv} %

\usepackage{algorithm}
\usepackage{algpseudocode}
\usepackage[utf8]{inputenc} %
\usepackage[T1]{fontenc}    %
\usepackage[pagebackref=true,breaklinks=true,letterpaper=true,colorlinks,bookmarks=false]{hyperref}
\usepackage{url}            %
\usepackage{booktabs}       %
\usepackage{amsmath}
\usepackage{amssymb}
\usepackage{nicefrac}
\usepackage{soul}        
\usepackage{lipsum}   %
\usepackage{wrapfig}
\usepackage{setspace}
\usepackage{xcolor}
\usepackage{flushend}
\usepackage{times}
\usepackage{epsfig}
\usepackage{graphicx}
\usepackage{dsfont}
\usepackage{dblfloatfix}
\usepackage{tikz}
\usepackage{pgfplots}
\usepackage{pgfplotstable}
\pgfplotsset{compat=newest}
\usepgfplotslibrary{external}

\usepackage[capitalize]{cleveref}
\crefname{section}{Sec.}{Secs.}
\Crefname{section}{Section}{Sections}
\Crefname{table}{Table}{Tables}
\crefname{table}{Tab.}{Tabs.}
\usetikzlibrary{patterns}
\usepackage{multirow}

\newcommand{\mycomment}[1]{}

\DeclareMathOperator*{\argmax}{arg\,max}

\DeclareMathOperator*{\topB}{top_B}

\newcommand{\comment} [1]{{\color{orange} \Comment     #1}} %

\def\nlsp{\hspace{-12pt}}

\def\nssp{\hspace{-3pt}}

\def\msp{\hspace{6pt}}

\newcommand{\equ}[1]{(\ref{equ:#1})\xspace}

\def\l1{\ensuremath{\ell_1}\xspace}
\def\l2{\ensuremath{\ell_2}\xspace}

\newcommand{\real}{\mathbb{R}}

\newcommand{\cL}{\mathcal{L}}

\newcommand{\cX}{\mathcal{X}}

\makeatletter
\DeclareRobustCommand\onedot{\futurelet\@let@token\@onedot}
\def\@onedot{\ifx\@let@token.\else.\null\fi\xspace}
 
\def\ie{\emph{i.e}\onedot} 
\def\vs{\emph{vs\onedot}}
 
 \def\vs{\emph{vs}\onedot}
 
\def\etal{\emph{et al}\onedot}
\makeatother

\begin{document}

\title{Training Ensembles with Inliers and Outliers \\for Semi-supervised Active Learning}

\author{Vladan Stojnić \hspace{10ex} Zakaria Laskar \hspace{10ex} Giorgos Tolias\\
Visual Recognition Group, Faculty of Electrical Engineering, Czech Technical University in Prague\\
{\tt\small stojnvla,laskazak,toliageo@fel.cvut.cz}
}
\maketitle
\maketitle

\begin{abstract}
Deep active learning in the presence of outlier examples poses a realistic yet challenging scenario. Acquiring unlabeled data for annotation requires a delicate balance between avoiding outliers to conserve the annotation budget and prioritizing useful inlier examples for effective training. In this work, we present an approach that leverages three highly synergistic components, which are identified as key ingredients: joint classifier training with inliers and outliers, semi-supervised learning through pseudo-labeling, and model ensembling.
Our work demonstrates that ensembling significantly enhances the accuracy of pseudo-labeling and improves the quality of data acquisition. By enabling semi-supervision through the joint training process, where outliers are properly handled, we observe a substantial boost in classifier accuracy through the use of all available unlabeled examples. Notably, we reveal that the integration of joint training renders explicit outlier detection unnecessary; a conventional component for acquisition in prior work.
The three key components align seamlessly with numerous existing approaches. Through empirical evaluations, we showcase that their combined use leads to a performance increase. Remarkably, despite its simplicity, our proposed approach outperforms all other methods in terms of performance. Code: \url{https://github.com/vladan-stojnic/active-outliers}

\end{abstract}

\section{Introduction}
\label{sec:intro}
Deep learning achieves considerable results on a variety of tasks but is data-hungry, while data annotation has a high cost and is a tedious process.
Using the annotation budget wisely is important, which is the focus of active learning~\cite{ren2021survey}.
Given a small labeled and a large unlabeled set, the goal is to acquire a subset of the latter to annotate via a labeling oracle. 
The acquisition function is responsible for selecting examples that will benefit the classifier training the most compared to the given labeled set. 
Acquisition, annotation, and model training are three steps that are typically interleaved over a sequence of consecutive rounds.

\begin{figure}[t]
  \begin{center}
  \includegraphics[width=0.47\textwidth]{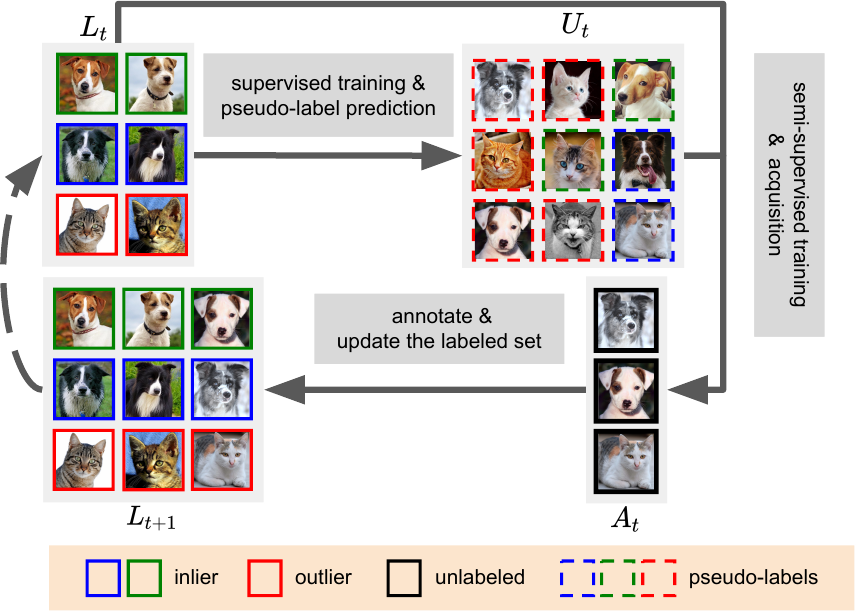}
  \end{center}
  \vspace{-15pt} %
  \caption{Overview of an active learning round with semi-supervision. $L_t$, $U_t$, and $A_t$ denote labeled,  unlabeled, and acquired sets at acquisition round $t$, respectively.
  \label{fig:teaser}
  \vspace{-15pt}
  }
\end{figure}

Early research is solely focusing on unlabeled sets that are outlier free, \ie all examples come from the categories of interest.
Information-theoretic criteria~\cite{entropy,bald} per example are often used to promote the most uncertain predictions for annotation, while other methods process all examples jointly and focus on diversity by pairwise comparisons~\cite{similar} or large coverage~\cite{coreset}. 
Surprisingly, random selection achieves good results~\cite{oriane,mittal2019parting} in deep active learning.
This is pronounced even more with the use of semi-supervision, which is deemed more important than the acquisition itself.

However, the outlier-free setup is not realistic. Nevertheless, the presence of outliers is a setup that has attracted less attention.
Avoiding outliers is essential so as not to waste the annotation budget.
Therefore, existing work~\cite{ccal,similar,lfosa,mqnet} explicitly or implicitly performs outlier detection and filtering.
Our proposed framework optionally includes outlier filtering, but we question its necessity and investigate the conditions under which it may be useful.
Additionally, standard acquisition functions are previously shown to fail~\cite{ccal,similar,lfosa,mqnet}, but we discover that simple ingredients are missing to make them competitive.

Acquiring outliers together with inliers is unavoidable, especially when their presence is extensive.
Therefore, we opt to take advantage of them by training a joint classifier for the inlier classes and the outlier class.
The joint classifier obtains increased inlier-class accuracy after each acquisition round and improved ability to perform acquisition.
Despite the fact that the joint training enables outlier detection and filtering, we show that, given an appropriate acquisition function, outlier filtering may not be necessary. 
The performance gap between the use of filtering and no filtering decreases for better acquisition functions.
The joint classifier has an additional benefit. It provides pseudo-labels for both types of examples allowing us to exploit semi-supervision.
Finally, we rely on the power of ensembles to improve pseudo-labeling and to equip the acquisition with a measure of statistical dispersion.
We do not use ensembles during testing, a choice that does not compromise the test-time complexity. An overview of the proposed approach is shown in Figure~\ref{fig:teaser}.

In summary, the proposed method consists of several components whose synergy is a key ingredient in surpassing all existing methods, while some of these components are applied for the first time to this specific setup.
Our findings are also useful in combination with other approaches and are expected to be useful for future approaches.
We perform experiments for a varying amount of outliers on benchmarks created from ImageNet, CIFAR100, and TinyImageNet datasets. 
The contributions on active learning of this work are summarized as follows: 
\begin{itemize}
\setlength\itemsep{-1pt}
\item We demonstrate the effectiveness of joint training with inliers and outliers, enabling the use of standard acquisition functions that were previously deemed ineffective in the presence of outliers.
\item  We introduce the use of semi-supervision as a key ingredient for achieving high performance, which is the first application of this technique in the context of outlier-inclusive scenarios.
\item  Our approach incorporates ensembles during training only, resulting in a significant performance boost without compromising test-time complexity.
\item The key components of this work are theoretically compatible with existing approaches; the practical performance benefits of their combination are empirically demonstrated in our experiments.
\item We conduct an extensive evaluation across a wide range of outlier percentages, from 0\% to 90\%, using non-tiny resolution images. Furthermore, we commit to sharing our code and experimental protocol publicly, aiming to enhance consistency in experimental setups across future studies.
\end{itemize}

\section{Related work}
\label{sec:related}
\looseness=-1
We review the related work on different setups of active learning, on the related task of semi-supervised learning with outliers, and on outlier detection in an open-world setting.
\subsection{Active learning}

\textbf{Acquisition in outlier-free active learning:} The two main families of scoring function are uncertainty-based and diversity-based. Uncertain examples are assumed informative, while uncertainty is measured in different ways, such as entropy~\cite{entropy}, confidence~\cite{conf}, margin~\cite{margin,base}, or mutual information between model parameters and model predictions~\cite{bald}. Improved uncertainty prediction is obtained through model ensembles~\cite{beluch2018power} or multiple input augmentations~\cite{hhk+20}. Other definitions of uncertainty use prediction inconsistency over input augmentations~\cite{gao2020consistency} or feature perturbations~\cite{parvaneh2022active}.
\looseness=-1
The second family includes methods that use the diversity of examples in the acquired set. CoreSet~\cite{coreset} and Cluster-Margin~\cite{citovsky2021batch} select diverse examples that well approximate the whole unlabeled set, while other works~\cite{cdg+21,azk+20} combine both notions of diversity and uncertainty.
Hacohen \etal~\cite{hacohen2022active} propose to annotate diverse but certain examples in low-budget regimes but diverse and uncertain examples in the high-budget regime. All these methods are developed for and evaluated in outlier-free setups but are known to fail with outliers~\cite{ccal,similar}.
In this work, we show under which conditions such approaches become effective again.

\textbf{Unlabeled examples in active learning:} Recent work~\cite{oriane, mittal2019parting, gao2020consistency, song2019combining, luth2023toward} demonstrates that the way unlabeled examples are used in learning is much more important than the selection process itself. This is the case for semi-supervision~\cite{oriane, mittal2019parting} and for self-supervision in the pre-training stage~\cite{oriane}. Different selection strategies make little or no difference in these setups, with random selection remaining a good enough choice, whose performance is often under-reported in the fully supervised setup~\cite{munjal2022towards}. Improper classifier and hyperparameter tuning lead to unfair method comparisons, requiring proper benchmarking~\cite{munjal2022towards,luth2023toward}. The aforementioned semi-supervised methods are not directly applicable to our setup due to the presence of outliers, which is one of the issues we handle in our work.
\looseness=-1
Other examples include the popular consistency criterion~\cite{tv17,bcg+19} that is used to perform the acquisition by Gao~\etal~\cite{gzy+20}, acquisition via classifiers trained to distinguish between labeled and unlabeled examples~\cite{vaal, tavaal}, and synthesizing examples with GANs~\cite{gan}.

\textbf{Outliers in active learning} are inherently present in real-world cases. 
Two recent methods, namely CCAL~\cite{ccal} and SIMILAR~\cite{similar}, propose to counterbalance informativeness and diversity with inlier confidence. 
CCAL~\cite{ccal} uses self-supervision in an innovative way to improve acquisition but discards it for classifier training, while we show that it is very beneficial for improving classification accuracy.
SIMILAR is the top-performing competitor, whose acquisition is elegant and principled but suffers from low scalability due to the costly optimization of each round. 
LfOSA~\cite{lfosa} filters out unlabelled examples based on outlier-class confidence and then performs the selection based on the maximum confidence.
MQNet~\cite{mqnet} presents acquisition as a purity-informativeness dilemma, meaning that a good acquisition function should balance purity, \ie the proportion of inlier samples in the acquired set, and informativeness. To construct such a function, they train an MLP on top of standard measures from the literature.
Assuming an unknown outlier percentage, these approaches should also work in the outlier-free setup. Nevertheless, we compare and show that this is not always the case.
Our method is simpler and scalable, demonstrates synergy with existing scoring functions, and enjoys the benefits of semi-supervision, which is a key ingredient for boosting performance.

\subsection{Outliers in semi-supervised learning}
Similarly to active learning with outliers, the goal is to minimize outlier influence, and all methods rely on different kinds of outlier detectors.
MTC~\cite{mtc}, D3SL~\cite{d3sl}, and UASD~\cite{uasd} rely on the Otsu threshold, SSL, and ensembles, respectively, to improve detection.
RETRIEVE~\cite{killamsetty2021retrieve} additionally proposes sub-modular functions to select a subset of good coverage.
OpenMatch~\cite{saito2021openmatch} uses one one-vs-all outlier detector per inlier class to overcome the lack of labeled outliers, which is a major difference from our setup, where an increasing amount of labeled outliers becomes available over consecutive rounds.

\subsection{Outlier detection}
Outlier detection or anomaly detection~\cite{gradnorm, csi, ssd, oodsurvey} is a relevant task that aims to solve a binary classification problem of correctly detecting outlier examples. Post-hoc outlier detection methods propose a scoring mechanism on top of an already trained feature backbone~\cite{mahalaood} to detect outlier examples. Some scoring measures include distance in the feature space~\cite{mahalaood}, pseudo-label confidence~\cite{lls18}, or entropy~\cite{gradnorm}. Other approaches~\cite{csi, ssd, hendrycks2018deep, bevandic2018discriminative, malinin2018predictive, lee2017training} adjust the training to maximize the test-time separability of inliers and outliers.
The main challenge is the unavailability of outliers during training. This is based on the assumption that outliers can come from any distribution in an open-set setup, and thus the objective is to learn an unbiased detector. This is not true in our active learning setup, where we have access to both unlabeled and labeled outliers; the latter typically become available after the first acquisition round.

\section{Method}
\label{sec:method}
We define the task of active learning with multiple acquisition rounds and present the proposed approach.

\begin{figure*}
\begin{center}
\includegraphics[width=.96\textwidth]{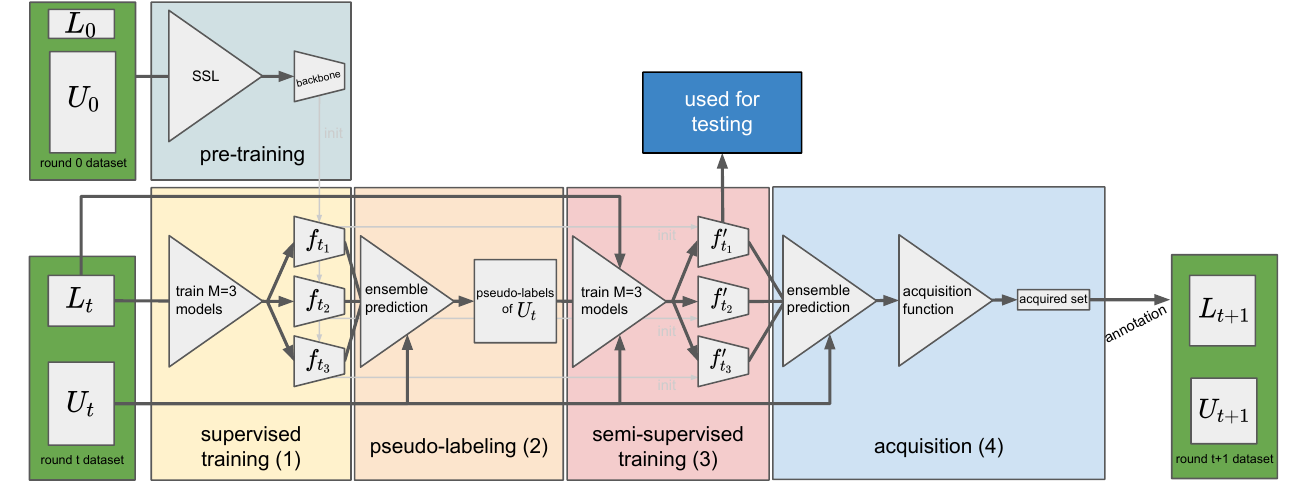}
\end{center}
\caption{Overview of all active learning stages for the proposed approach during round $t$, for $t>0$. It includes independently training $M$ networks for $K+1$-way classification, acquisition, and annotation. Acquisition exploits the ensemble classifier predictions on the unlabeled set and optionally includes outlier filtering. During round 0, SSL is employed to train the backbone, which is used as initialization for rounds $t>0$, and acquisition with random sampling is performed (not shown in the figure). Testing is performed for $K$-way classification on inliers only with a single network; ensembles are only used for internal processes, \ie pseudo-labeling and acquisition.\label{fig:overview}
\vspace{-3pt}
}
\end{figure*}

\subsection{Task formulation}
We consider active learning for the classification of object categories $C$, with $K=|C|$. 
We consider an additional class, called outlier class $C_o$, meant for examples that are not from inlier classes $C$. Examples from $C$ and $C_o$ are called inliers and outliers, respectively.
Initially, we are given a labeled set $L_0$ and an unlabeled set $U_0$, which consist of inliers only, and both inliers and outliers, respectively. 
Active learning consists of sequential rounds that include the \emph{acquisition} of a subset of the unlabeled set and the \emph{annotation} of the acquired subset.
The acquisition should satisfy two objectives that are challenging to balance. Firstly, acquired sets should be as outlier-free as possible. Secondly, the newly acquired and annotated examples should contribute the most to improving the $K$-way classifier compared to the current labeled set.

At round $t$, the acquisition process makes use of the available examples in $L_t$ and $U_t$
to select set $A_t \subset U_{t}$ whose labels are assigned by an annotation oracle. 
Acquired set $A_t$ may include outliers too, which are labeled by class $C_o$.
Then, the two sets are updated, \ie $L_{t+1} = L_{t} \cup A_{t}$ and $U_{t+1} = U_{t} \setminus A_{t}$. The size of $A_{t}$ is fixed and equal to the annotation budget per round $B = |A_{t}|$, which is a parameter of the task.

\subsection{Overview}
\label{sec:overview}
We start by training a backbone network on all examples in $L_0 \cup U_0$ by Self-Supervised Learning (SSL), which is known to be beneficial for active learning~\cite{oriane}. 
At the beginning of round $t$ for $t>0$, $M$ deep network classifiers are initialized by the result of SSL and trained for $K+1$ classes with examples in $L_t$.
The independently trained classifiers are ensembled to perform pseudo-labeling of $U_t$, which are used to continue training the $M$ classifiers in a semi-supervised way on the union of $L_t$ and $U_t$.
Then, the ensemble of classifiers, trained with semi-supervision, is used to equip the acquisition process, which optionally includes explicit outlier filtering along with a measure of example uncertainty and/or diversity. 
Acquired examples are finally annotated by a labeling oracle to obtain $L_{t+1}$ and $U_{t+1}$.
Round 0 is a special case that we discuss towards the end of this section.

During test time we do not use any ensembles, which are only used during internal processes to improve pseudo-labeling and acquisition. The test accuracy is evaluated with a single network which is the result of semi-supervised learning.
The overall process is summarized in Figure~\ref{fig:overview}, depicting all stages of a single round. 

\subsection{Training} 
\label{sec:classifier}
The network classifier is a function $f: \cX \rightarrow \real^{K+1}$, where $\cX$ is the space of all examples, and the output space consists of all inlier classes and the outlier one. 
We consider $M$ different networks, and the predicted probability distribution for example $x\in \cX$ at round $t$ by network $i\in [1,\ldots,M]$ is denoted by $f_{t_i}(x)$.
Network ensembling is performed by averaging the $M$ output probabilities and is denoted by $F_t(x)$, and the probability of the $j$-th class is given by $F_{t}(x)_j$.
\looseness=-1

\paragraph{SSL pre-training:} Before the first round, SSL is performed by instance discrimination, where a positive pair is formed by two different augmentations of the same example, and a negative example is formed by simply picking a different example. This step uses all examples in $L_0 \cup U_0$ without any labels. SimCLR~\cite{simclr} is the method we choose, following the work of Du~\etal~\cite{ccal}. %
\looseness=-1

\paragraph{Supervised training: }
At the beginning of round $t$, each of the $M$ classifiers is trained by minimizing empirical loss 
\begin{equation}
\cL(L_t) = \frac{1}{|L_t|}\sum_{x \in L_t} \ell(f_{t_i}(x), y(x)), 
\label{equ:loss_full}
\end{equation}
where $y(x)\in [1,\ldots, K+1]$ is the label of $x$, and $\ell(\cdot)$ is the cross-entropy loss.

\paragraph{Semi-supervised training:} 
We use the unlabeled examples, but only after we first train in a fully supervised way with \equ{loss_full}.
Then, we generate pseudo-labels $\hat{y}_t(x)=\argmax_j F_t(x)_j$ for all examples in $U_t$. 
Each pseudo-label is assigned a weight according to the certainty of the prediction given by
\begin{equation}
w_t(x) = 1 - \frac{H\left(F_{t}(x)\right)}{\log (K+1)},
\label{equ:weight}
\end{equation}
which is inversely proportional to the normalized entropy and bounded in $[0,1]$, with entropy given by function $H$. 
We initialize $M$ classifiers with the result obtained by \equ{loss_full}, but use both labeled and unlabeled examples with weighted loss terms given by 
\begin{equation}
\cL_\text{semi}(L_t, U_t) = \frac{1}{N}\sum_{x \in L_t \cup U_t} w_t(x) \ell(f^\prime_{t_i}(x), \hat{y}_t(x)), 
\label{equ:loss_semi}
\end{equation}
where $\hat{y}_t(x)=y(x)$ and $w_t(x) = 1$ for the labeled examples, and $N=|L_t|+|U_t|$. 
We use $f^\prime_{t_i}$ and $F^\prime_t$ for the networks obtained with this semi-supervised way to differentiate from the ones of the previous stage.
To evaluate the classification accuracy of round $t$, one of the $M$ networks is randomly picked and used.

\paragraph{Round 0:} 
Before any acquisition, at round $t=0$, there are no labeled outliers; therefore, training the $K+1$-way classifier is not possible. We simply perform random acquisition at this stage. 
In summary, we train the backbone via SSL and then perform random acquisition and annotation that results in $L_1$ and $U_1$.

\subsection{Acquisition}
To acquire examples during round $t$, we exploit prediction $F^\prime_{t}(x)$ for $x\in U_t$ to assign acquisition value $a_{t}(x)$.
This value is composed of a measure of example uncertainty or diversity, obtained via function $\tilde{a}_{t}(x)$, which is optionally combined with outlier filtering via the ensemble predictions.
In particular, the final acquisition value is $a_{t}(x) = \tilde{a}_{t}(x)\mathds{1}_{\hat{y}_t(x)\neq C_o}$ to include outlier filtering and assign 0 value to examples predicted as outliers, or just $a_{t}(x)=\tilde{a}_{t}(x)$ without filtering.
One of the standard choices is to rely on the ensemble of classifiers, measure statistical dispersion among them, and choose examples with large disagreement.
In particular, we estimate the Variation-Ratio (VR)~\cite{freeman1965elementary} given by
\begin{equation}
\tilde{a}_{t}(x) = 1-\frac{\left|\left\{i:~\hat{y}^{\prime}_{t_i}(x) =  \hat{y}^{\prime}_t(x) \right\}\right|}{M},
  \label{equ:acq}
\end{equation}
where $\hat{y}^{\prime}_{t_i} = \argmax_j f^\prime_{t_i}(x)_j$ is the pseudo-label of the $i$-th classifier. 
VR measures the proportion of pseudo-labels from a single classifier that disagree with the pseudo-label from the ensemble classifier. 
Examples with large disagreement get assigned large scores.
At the end, we sort examples in descending order based on $a_{t}(x)$ and acquire the first $B$ examples. 
Other candidate functions are entropy $\tilde{a}_{t}(x)=H(F_t(x))$, uniform random score generator $\tilde{a}_{t}(x) \sim \mathcal{U}_{[0,1]}$, CoreSet, maximum confidence $\tilde{a}_{t}(x)=1-\max_j F_t(x)_j$, and more.
In our experiments, we identify cases where outlier filtering is needed or not, and where simple measures become effective despite the presence of outliers.

\begin{figure*}[t]
  \vspace{-0pt} %
  \centering
  \input{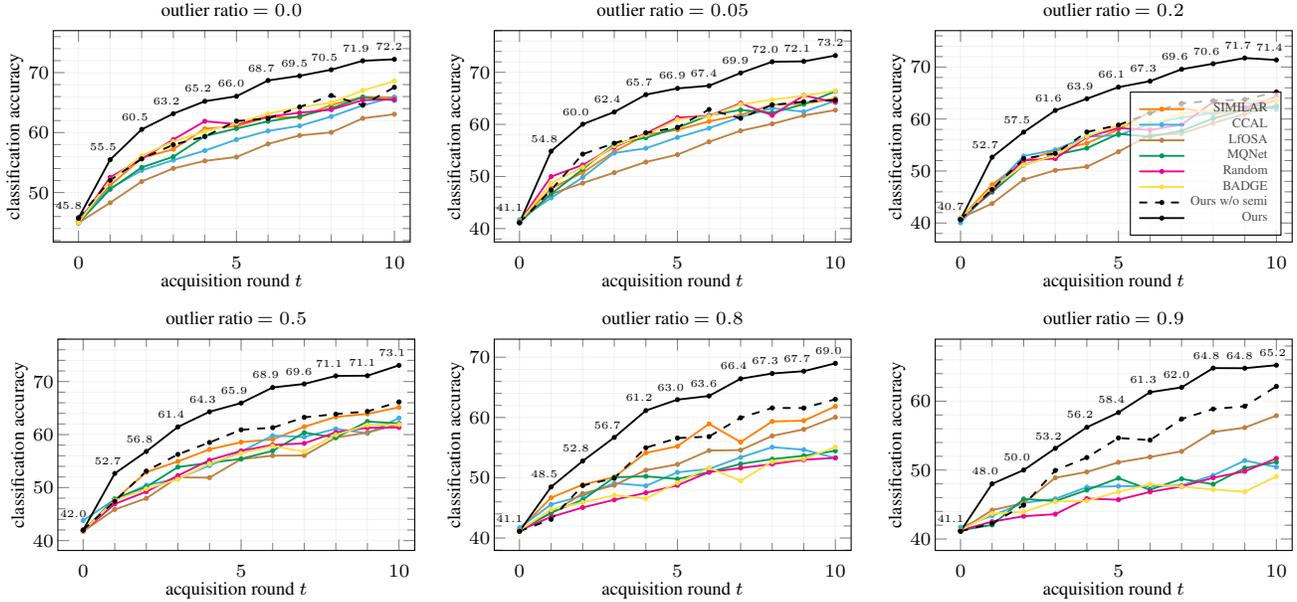}
  \vspace{-7pt} %
  \caption{Comparison of classification accuracy over multiple active learning rounds for varying outlier ratios on ImageNet. SIMILAR is excluded for 0.9 outlier ratio since we were not able to run it even on a machine with 800GB of RAM.
  \label{fig:main_fig_acc}
  \vspace{-0pt}
  }
\end{figure*}

\begin{figure}[b]
  \vspace{-0pt} %
  \centering
  \input{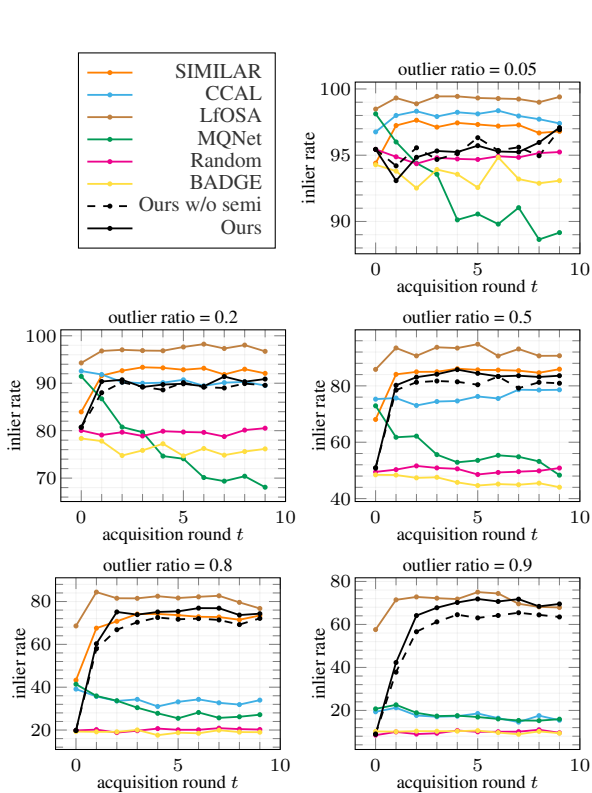}
\pgfplotsset{every tick label/.append style={font=\scriptsize}}
\pgfplotsset{select coords between index/.style 2 args={
    x filter/.code={
        \ifnum\coordindex<#1\def\pgfmathresult{}\fi
        \ifnum\coordindex>#2\def\pgfmathresult{}\fi
    }
}}
\pgfplotsset{minor grid style={solid,gray,opacity=0.1}}
\pgfplotsset{major grid style={solid,gray,opacity=0.1}}
\vspace{-10pt}
\begin{tabular}{@{\msp}cc@{\msp}} 
\multicolumn{2}{c}{
\begin{tikzpicture}
\hspace{15pt}\begin{axis}[%
	width=130pt,
	height=110pt,
	xlabel={\scriptsize acquisition round $t$},
	ylabel={\scriptsize inlier rate},
	ylabel style={yshift=-5pt},
    y tick label style={xshift=2pt},
    xtick={0,5,...,10},
    xmax = 10,
    xmin = -1,
    legend pos= {south east},
    title style={align=center, font=\scriptsize, row sep=3pt},
	title={\scriptsize{outlier ratio = 0.05}},
	title style={yshift=-7pt},    
	xlabel style={yshift=5pt},    
    legend style={cells={anchor=east}, font =\footnotesize, fill opacity=0.8, row sep=-2.5pt, xshift=-25ex},
    grid=both,
    minor tick num=4,
]
	\addplot[color=orange,    mark=*, mark size=0.6, line width=0.7, select coords between index={0}{10}] table[x=al, y=similar] \idrinfiveood;\addlegendentry{SIMILAR};
	 \addplot[color=CornflowerBlue,    mark=*, mark size=0.6, line width=0.7, select coords between index={0}{10}] table[x=al, y=ccal] \idrinfiveood;\addlegendentry{CCAL};
    \addplot[color=brown,    mark=*, mark size=0.6, line width=0.7, select coords between index={0}{10}] table[x=al, y=lfosa] \idrinfiveood;\addlegendentry{LfOSA};
    \addplot[color=ForestGreen,    mark=*, mark size=0.6, line width=0.7, select coords between index={0}{10}] table[x=al, y=mqnet] \idrinfiveood;\addlegendentry{MQNet};
	\addplot[color=Magenta,    mark=*, mark size=0.6, line width=0.7, select coords between index={0}{10}] table[x=al, y=random] \idrinfiveood;\addlegendentry{Random};
	\addplot[color=Goldenrod,    mark=*, mark size=0.6, line width=0.7, select coords between index={0}{10}] table[x=al, y=badge] \idrinfiveood;\addlegendentry{BADGE};
	\addplot[color=black,     dashed, mark=*, mark options={solid},  mark size=0.6, line width=0.7, select coords between index={0}{10}] table[x=al, y=ours_fully] \idrinfiveood;\addlegendentry{Ours w/o semi};
	\addplot[color=black,    mark=*, mark size=0.6, line width=0.7, select coords between index={0}{10}] table[x=al, y=ours_semi] \idrinfiveood;\addlegendentry{Ours};
\end{axis}
\end{tikzpicture}
}
\hspace{-15pt}
\\[-5pt]
\begin{tikzpicture}
\begin{axis}[%
	width=130pt,
	height=110pt,
	xlabel={\scriptsize acquisition round $t$},
	ylabel={\scriptsize inlier rate},
	ylabel style={yshift=-5pt},
    y tick label style={xshift=2pt},
    xtick={0,5,...,10},
    xmax = 10,
    xmin = -1,
    legend pos= {south east},
    legend style={cells={anchor=east}, font =\scriptsize, fill opacity=0.8, row sep=-2.5pt},
    title style={align=center, font=\scriptsize, row sep=3pt},
	title={\scriptsize{outlier ratio = 0.2}},
	title style={yshift=-7pt},    
	xlabel style={yshift=5pt},    
    grid=both,
    minor tick num=4,
]
	\addplot[color=orange,    mark=*, mark size=0.6, line width=0.7, select coords between index={0}{10}] table[x=al, y=similar] \idrintwentyood;%
	 \addplot[color=CornflowerBlue,    mark=*, mark size=0.6, line width=0.7, select coords between index={0}{10}] table[x=al, y=ccal] \idrintwentyood;%
    \addplot[color=brown,    mark=*, mark size=0.6, line width=0.7, select coords between index={0}{10}] table[x=al, y=lfosa] \idrintwentyood;%
    \addplot[color=ForestGreen,    mark=*, mark size=0.6, line width=0.7, select coords between index={0}{10}] table[x=al, y=mqnet] \idrintwentyood;%
	\addplot[color=Magenta,    mark=*, mark size=0.6, line width=0.7, select coords between index={0}{10}] table[x=al, y=random] \idrintwentyood;%
	\addplot[color=Goldenrod,    mark=*, mark size=0.6, line width=0.7, select coords between index={0}{10}] table[x=al, y=badge] \idrintwentyood;%
	\addplot[color=black,     dashed, mark=*, mark options={solid},  mark size=0.6, line width=0.7, select coords between index={0}{20}] table[x=al, y=ours_fully] \idrintwentyood;%
	\addplot[color=black,    mark=*, mark size=0.6, line width=0.7, select coords between index={0}{20}] table[x=al, y=ours_semi] \idrintwentyood;%
\end{axis}
\end{tikzpicture}
\hspace{-15pt}
&
\begin{tikzpicture}
\begin{axis}[%
	width=130pt,
	height=110pt,
	xlabel={\scriptsize acquisition round $t$},
	ylabel={\scriptsize inlier rate},
	ylabel style={yshift=-5pt},
    y tick label style={xshift=2pt},
    xtick={0,5,...,10},
    xmax = 10,
    xmin = -1,
    legend pos= {south east},
    legend style={cells={anchor=east}, font =\scriptsize, fill opacity=0.8, row sep=-2.5pt},
    title style={align=center, font=\scriptsize, row sep=3pt},
	title={\scriptsize{outlier ratio = 0.5}},
	title style={yshift=-7pt},    
	xlabel style={yshift=5pt},    
    grid=both,
    minor tick num=4,
]
	\addplot[color=orange,    mark=*, mark size=0.6, line width=0.7, select coords between index={0}{10}] table[x=al, y=similar] \idrinfiftyood;
	 \addplot[color=CornflowerBlue,    mark=*, mark size=0.6, line width=0.7, select coords between index={0}{10}] table[x=al, y=ccal] \idrinfiftyood;
    \addplot[color=brown,    mark=*, mark size=0.6, line width=0.7, select coords between index={0}{10}] table[x=al, y=lfosa] \idrinfiftyood;
    \addplot[color=ForestGreen,    mark=*, mark size=0.6, line width=0.7, select coords between index={0}{10}] table[x=al, y=mqnet] \idrinfiftyood;
	\addplot[color=Magenta,    mark=*, mark size=0.6, line width=0.7, select coords between index={0}{10}] table[x=al, y=random] \idrinfiftyood;
	\addplot[color=Goldenrod,    mark=*, mark size=0.6, line width=0.7, select coords between index={0}{10}] table[x=al, y=badge] \idrinfiftyood;
	\addplot[color=black,     dashed, mark=*, mark options={solid},  mark size=0.6, line width=0.7, select coords between index={0}{20}] table[x=al, y=ours_fully] \idrinfiftyood;
	\addplot[color=black,    mark=*, mark size=0.6, line width=0.7, select coords between index={0}{20}] table[x=al, y=ours_semi] \idrinfiftyood;
\end{axis}
\end{tikzpicture}
\hspace{-15pt}
\\[-5pt]
\begin{tikzpicture}
\begin{axis}[%
	width=130pt,
	height=110pt,
	xlabel={\scriptsize acquisition round $t$},
	ylabel={\scriptsize inlier rate},
	ylabel style={yshift=-5pt},
    y tick label style={xshift=2pt},
    xtick={0,5,...,10},
    xmax = 10,
    xmin = -1,
    legend pos= {south east},
    legend style={cells={anchor=east}, font =\scriptsize, fill opacity=0.8, row sep=-2.5pt},
    title style={align=center, font=\scriptsize, row sep=3pt},
	title={\scriptsize{outlier ratio = 0.8}},
	title style={yshift=-7pt},    
	xlabel style={yshift=5pt},    
    grid=both,
    minor tick num=4,
]
	\addplot[color=orange,    mark=*, mark size=0.6, line width=0.7, select coords between index={0}{10}] table[x=al, y=similar] \idrineightyood;
	 \addplot[color=CornflowerBlue,    mark=*, mark size=0.6, line width=0.7, select coords between index={0}{10}] table[x=al, y=ccal] \idrineightyood;
    \addplot[color=brown,    mark=*, mark size=0.6, line width=0.7, select coords between index={0}{10}] table[x=al, y=lfosa] \idrineightyood;
    \addplot[color=ForestGreen,    mark=*, mark size=0.6, line width=0.7, select coords between index={0}{10}] table[x=al, y=mqnet] \idrineightyood;
	\addplot[color=Magenta,    mark=*, mark size=0.6, line width=0.7, select coords between index={0}{10}] table[x=al, y=random] \idrineightyood;
	\addplot[color=Goldenrod,    mark=*, mark size=0.6, line width=0.7, select coords between index={0}{10}] table[x=al, y=badge] \idrineightyood;
	\addplot[color=black,     dashed, mark=*, mark options={solid},  mark size=0.6, line width=0.7, select coords between index={0}{20}] table[x=al, y=ours_fully] \idrineightyood;
	\addplot[color=black,    mark=*, mark size=0.6, line width=0.7, select coords between index={0}{20}] table[x=al, y=ours_semi] \idrineightyood;
\end{axis}
\end{tikzpicture}
\hspace{-15pt}
&
\begin{tikzpicture}
\begin{axis}[%
	width=130pt,
	height=110pt,
	xlabel={\scriptsize acquisition round $t$},
	ylabel={\scriptsize inlier rate},
	ylabel style={yshift=-5pt},
    y tick label style={xshift=2pt},
    xtick={0,5,...,10},
    xmax = 10,
    xmin = -1,
    legend pos= {south east},
    legend style={cells={anchor=east}, font =\scriptsize, fill opacity=0.8, row sep=-2.5pt},
    title style={align=center, font=\scriptsize, row sep=3pt},
	title={\scriptsize{outlier ratio = 0.9}},
	title style={yshift=-7pt},    
	xlabel style={yshift=5pt},    
    grid=both,
    minor tick num=4,
]
	 \addplot[color=CornflowerBlue,    mark=*, mark size=0.6, line width=0.7, select coords between index={0}{10}] table[x=al, y=ccal] \idrinninetyood;
    \addplot[color=brown,    mark=*, mark size=0.6, line width=0.7, select coords between index={0}{10}] table[x=al, y=lfosa] \idrinninetyood;
    \addplot[color=ForestGreen,    mark=*, mark size=0.6, line width=0.7, select coords between index={0}{10}] table[x=al, y=mqnet] \idrinninetyood;
	\addplot[color=Magenta,    mark=*, mark size=0.6, line width=0.7, select coords between index={0}{10}] table[x=al, y=random] \idrinninetyood;
	\addplot[color=Goldenrod,    mark=*, mark size=0.6, line width=0.7, select coords between index={0}{10}] table[x=al, y=badge] \idrinninetyood;
	\addplot[color=black,     dashed, mark=*, mark options={solid},  mark size=0.6, line width=0.7, select coords between index={0}{10}] table[x=al, y=ours_fully] \idrinninetyood;
	\addplot[color=black,    mark=*, mark size=0.6, line width=0.7, select coords between index={0}{10}] table[x=al, y=ours_semi] \idrinninetyood;
\end{axis}
\end{tikzpicture}
\hspace{-15pt}
\end{tabular}
  \vspace{-0pt} %
  \caption{
  Comparison of inlier rate over multiple active learning rounds for varying outlier ratios on ImageNet. Round 0 is with random selection for our approach and with their specific choice for each method. Reporting round 10 is not included because testing and evaluation are performed before acquisition.
  \label{fig:main_fig_idr}
\vspace{-0pt} %
  }
\end{figure}

\section{Experiments}
\label{sec:exp}
We discuss training details, datasets, experimental protocol, competing methods, and present the results. 

\textbf{Datasets and experimental setup: }
\label{sec:datasets}
Most existing methods do not share experimental setup details~\cite{ccal}, release source code~\cite{lfosa}, evaluate on a wide range of outlier ratios~\cite{similar}, or evaluate on non-tiny resolution images~\cite{similar,ccal,lfosa}. To address this, we use ImageNet ILSVRC2012~\cite{rds+15}, TinyImageNet~\cite{tiny}, and CIFAR100~\cite{cifar}, to generate benchmarks for active learning with outliers.

The amount of outlier examples in the initial unlabeled set is quantified by the \emph{outlier ratio}, \ie the ratio of the number of outliers over the number of all examples.
We consider 25 ImageNet classes that correspond to dog breeds as inlier classes and examples from 700 different classes as outliers. 
We use the training split to create $L_0$ and $U_0$. In particular, $L_0$ is generated by 20  randomly chosen examples per inlier class, and $U_0$ by 500 randomly chosen examples per inlier class and randomly chosen outlier examples so that the outlier ratio is equivalent to 0, 0.05, 0.2, 0.5, 0.8, 0.9 for six different benchmarks. The outlier examples of a particular outlier ratio are a subset of the outlier examples for a larger outlier ratio.
The test set is formed by examples from the validation split and contains 1,250 examples from the inlier classes.
In a similar way, we generate benchmarks from TinyImageNet and CIFAR100. 
We report average classification accuracy over 5 seeds defining different selections for $L_0$, but always the same $U_0$.

Accuracy reported for round $t$ is with the classifier trained after $t$ acquisition rounds, \ie one for round 0 and $t-1$ for the follow-up rounds, but the acquisition of round $t$ is not included. As a reference comparison, we train classifiers with \equ{loss_full} at $t=0$ and report the achieved classification accuracy. 
We report the acquisition \emph{inlier rate}, which is the percentage of inliers among the acquired set per round, defined for round $0$ with random acquisition and for all other rounds except the last one, where testing is performed before acquisition. 
The annotation budget is $B=500$ for ImageNet and $B=100$ for TinyImageNet and CIFAR100.
We use ResNet18~\cite{hzr+16} as the backbone.
Details regarding CIFAR100 and TinyImageNet benchmarks, detailed reporting of average performance and standard deviation in table format for our main experiments, and additional implementation details can be found in the appendix.
Additionally, we evaluate our method on the experimental setup followed by MQNet~\cite{mqnet} and report results in the appendix.

\textbf{Baselines and other methods: }
The main variant of \textbf{our approach} is with $M=5$, includes outlier filtering, and uses VR as a scoring function, unless otherwise stated. We often refer to the numbered steps, as shown in Figure~\ref{fig:overview}, to clarify particular design choices in ablations.
The following acquisition functions are used within our approach: \textbf{Random} selection, \textbf{Entropy}-based selection~\cite{entropy}, \textbf{VR}~\cite{freeman1965elementary}, and \textbf{CoreSet}~\cite{coreset}.
We compare with \textbf{BADGE}~\cite{azk+20} as a recent, well-performing, and representative method developed for the outlier-free setup.
Additionally, we compare with \textbf{CCAL}~\cite{ccal}, \textbf{SIMILAR}~\cite{similar}, \textbf{LfOSA}~\cite{lfosa}, and \textbf{MQNet}~\cite{mqnet} as approaches that perform in the presence of outliers.
We observe that SSL-based network initialization is beneficial for all these methods; therefore, we use it to evaluate them. Due to this choice, the reported performance for these methods is noticeably higher than their off-the-shelf application.
Note that CCAL, in the original work, uses SSL for the backbones used in the acquisition process but not as classifier initialization.
We run CoreSet, BADGE, CCAL, SIMILAR, and MQNet using the provided implementations, after integrating them into our implementation framework. We implement LfOSA by ourselves.

\begin{figure}[t]
  \centering
  \vspace{-5pt} %
  \input{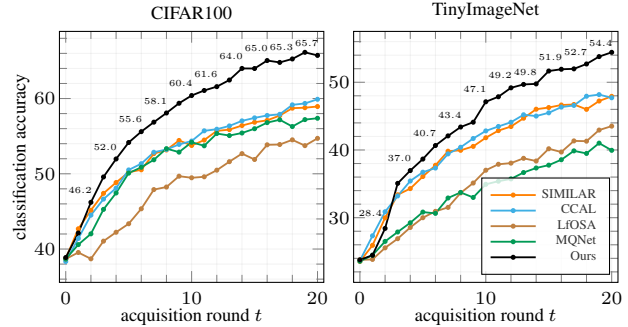}
\pgfplotsset{every tick label/.append style={font=\scriptsize}}
\pgfplotsset{select coords between index/.style 2 args={
    x filter/.code={
        \ifnum\coordindex<#1\def\pgfmathresult{}\fi
        \ifnum\coordindex>#2\def\pgfmathresult{}\fi
    }
}}
\pgfplotsset{minor grid style={solid,gray,opacity=0.1}}
\pgfplotsset{major grid style={solid,gray,opacity=0.1}}
\hspace{-15pt}
\begin{tabular}{@{\msp}c@{\msp}c}
\begin{tikzpicture}
\begin{axis}[%
	width=145pt,
	height=140pt,
	xlabel={\scriptsize acquisition round $t$},
	ylabel={\scriptsize classification accuracy},
	ylabel style={yshift=-5pt},
    y tick label style={xshift=2pt},
    xtick={0,10,...,20},
    xmax = 20.5,
    xmin = -0.5,
    legend pos= {south east},
    title style={align=center, font=\scriptsize, row sep=3pt},
        title={\scriptsize{CIFAR100}},
	title style={yshift=-5pt},    
	xlabel style={yshift=5pt},    
    grid=both,
    minor tick num=4,
]
	\addplot[color=orange,solid, mark=*,   mark size=0.6, line width=0.7, select coords between index={0}{20}] table[x=al, y=similar] \cifareightyood;
	\addplot[color=CornflowerBlue,solid, mark=*,   mark size=0.6, line width=0.7, select coords between index={0}{20}] table[x=al, y=ccal] \cifareightyood;
    \addplot[color=brown,solid, mark=*,   mark size=0.6, line width=0.7, select coords between index={0}{20}] table[x=al, y=lfosa] \cifareightyood;
    \addplot[color=ForestGreen,solid, mark=*,   mark size=0.6, line width=0.7, select coords between index={0}{20}] table[x=al, y=mqnet] \cifareightyood;
	\addplot[color=black,solid, mark=*,   mark size=0.6, line width=0.7, select coords between index={0}{20}] table[x=al, y=ours_semi] \cifareightyood;
    \node [above left] at (axis cs:  2.8,  46.21) {\tiny \textcolor{black}{\scalebox{.7}{$46.2$}}};
    \node [above left] at (axis cs:  4.8,  51.97) {\tiny \textcolor{black}{\scalebox{.7}{$52.0$}}};
    \node [above left] at (axis cs:  6.8,  55.60) {\tiny \textcolor{black}{\scalebox{.7}{$55.6$}}};
    \node [above left] at (axis cs:  8.8,  58.08) {\tiny \textcolor{black}{\scalebox{.7}{$58.1$}}};
    \node [above left] at (axis cs:  10.8,  60.40) {\tiny \textcolor{black}{\scalebox{.7}{$60.4$}}};
    \node [above left] at (axis cs:  12.8,  61.58) {\tiny \textcolor{black}{\scalebox{.7}{$61.6$}}};
    \node [above left] at (axis cs:  14.8,  64.00) {\tiny \textcolor{black}{\scalebox{.7}{$64.0$}}};
    \node [above left] at (axis cs:  16.8,  65.05) {\tiny \textcolor{black}{\scalebox{.7}{$65.0$}}};
    \node [above left] at (axis cs:  18.8,  65.26) {\tiny \textcolor{black}{\scalebox{.7}{$65.3$}}};
    \node [above left] at (axis cs:  20.8,  65.73) {\tiny \textcolor{black}{\scalebox{.7}{$65.7$}}};

\end{axis}
\end{tikzpicture}
\hspace{-15pt}
&
\begin{tikzpicture}
\begin{axis}[%
	width=145pt,
	height=140pt,
	xlabel={\scriptsize acquisition round $t$},
    y tick label style={xshift=2pt},
    xtick={0,10,...,20},
    xmax = 20.5,
    xmin = -0.5,
    legend pos= {south east},
    title style={align=center, font=\scriptsize, row sep=3pt},
        title={\scriptsize{TinyImageNet}},
	title style={yshift=-5pt},    
	xlabel style={yshift=5pt},    
    legend style={cells={anchor=east}, font =\tiny, fill opacity=0.8, row sep=-3pt},
    grid=both,
    minor tick num=4,
]
	\addplot[color=orange,solid, mark=*,   mark size=0.6, line width=0.7, select coords between index={0}{20}] table[x=al, y=similar] \tinyeightyood;\addlegendentry{SIMILAR};
	\addplot[color=CornflowerBlue,solid, mark=*,   mark size=0.6, line width=0.7, select coords between index={0}{20}] table[x=al, y=ccal] \tinyeightyood;\addlegendentry{CCAL};
    \addplot[color=brown,solid, mark=*,   mark size=0.6, line width=0.7, select coords between index={0}{20}] table[x=al, y=lfosa] \tinyeightyood;\addlegendentry{LfOSA};
    \addplot[color=ForestGreen,solid, mark=*,   mark size=0.6, line width=0.7, select coords between index={0}{20}] table[x=al, y=mqnet] \tinyeightyood;\addlegendentry{MQNet};
	\addplot[color=black,solid, mark=*,   mark size=0.6, line width=0.7, select coords between index={0}{20}] table[x=al, y=ours_semi] \tinyeightyood;\addlegendentry{Ours};
    \node [above left] at (axis cs:  2.5,  28.82) {\tiny \textcolor{black}{\scalebox{.7}{$28.4$}}};
    \node [above left] at (axis cs:  4.8,  36.96) {\tiny \textcolor{black}{\scalebox{.7}{$37.0$}}};
    \node [above left] at (axis cs:  6.8,  40.66) {\tiny \textcolor{black}{\scalebox{.7}{$40.7$}}};
    \node [above left] at (axis cs:  8.8,  43.39) {\tiny \textcolor{black}{\scalebox{.7}{$43.4$}}};
    \node [above left] at (axis cs:  10.8,  47.12) {\tiny \textcolor{black}{\scalebox{.7}{$47.1$}}};
    \node [above left] at (axis cs:  12.8,  49.18) {\tiny \textcolor{black}{\scalebox{.7}{$49.2$}}};
    \node [above left] at (axis cs:  14.8,  49.76) {\tiny \textcolor{black}{\scalebox{.7}{$49.8$}}};
    \node [above left] at (axis cs:  16.8,  51.92) {\tiny \textcolor{black}{\scalebox{.7}{$51.9$}}};
    \node [above left] at (axis cs:  18.8,  52.70) {\tiny \textcolor{black}{\scalebox{.7}{$52.7$}}};
    \node [above left] at (axis cs:  20.8,  54.0) {\tiny \textcolor{black}{\scalebox{.7}{$54.4$}}};

\end{axis}
\end{tikzpicture}
\end{tabular}
  \vspace{-12pt} %
  \caption{Comparison of classification accuracy over active learning rounds on CIFAR100 and TinyImageNet for 0.8 outlier ratio.
  \vspace{-15pt} %
   \label{fig:cifar_tiny}
   }
\end{figure}

\begin{figure*}[b!]
  \centering
  \vspace{-5pt}
  \input{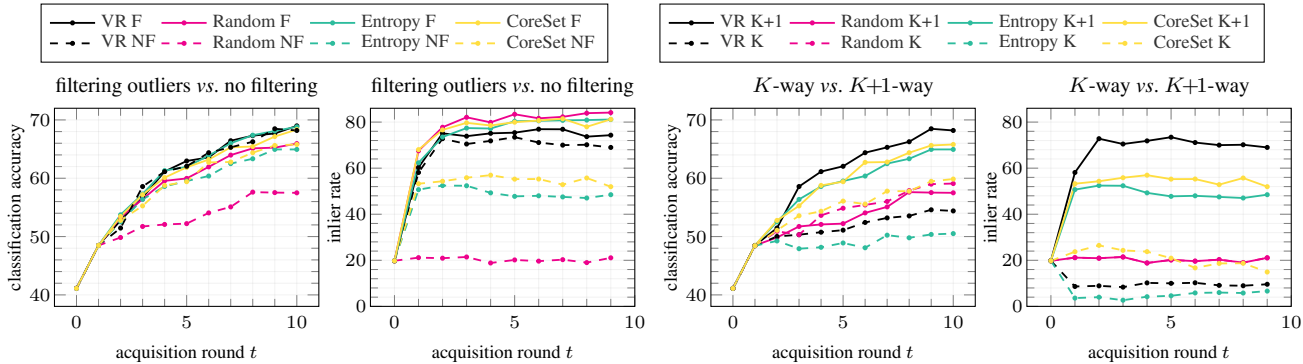}
\pgfplotsset{every tick label/.append style={font=\scriptsize}}
\pgfplotsset{select coords between index/.style 2 args={
    x filter/.code={
        \ifnum\coordindex<#1\def\pgfmathresult{}\fi
        \ifnum\coordindex>#2\def\pgfmathresult{}\fi
    }
}}
\pgfplotsset{minor grid style={solid,gray,opacity=0.1}}
\pgfplotsset{major grid style={solid,gray,opacity=0.1}}
\begin{tabular}{@{\nssp}c@{\nlsp}c@{\nssp}c@{\nlsp}c}
\begin{tikzpicture}
\begin{axis}[%
	width=145pt,
	height=120pt,
	xlabel={\scriptsize acquisition round $t$},
	ylabel={\scriptsize classification accuracy},
	ylabel style={yshift=-5pt},
    y tick label style={xshift=2pt},
    xtick={0,5,...,10},
    xmax = 11,
    xmin = -1,
    ymax = 72,
    legend columns = 4, legend pos= {south east},
    legend style={cells={anchor=west}, font =\scriptsize, fill opacity=0.8, row sep=-2.5pt, xshift=25ex, yshift=20ex},
    title style={align=center},
	title={\footnotesize filtering outliers \vs no filtering},
	title style={yshift=-5pt},    
	xlabel style={yshift=0pt},    
    grid=both,
    minor tick num=4,    
]
    \addplot[color=black,     solid, mark options={solid}, mark=*,   mark size=0.6, line width=0.7, select coords between index={0}{10}] table[x=al, y=vr_filt_k1] \filtablacc;\addlegendentry{VR F};
    \addplot[Magenta,     solid, mark options={solid}, mark=*,   mark size=0.6, line width=0.7, select coords between index={0}{10}] table[x=al, y=random_filt_k1] \filtablacc;\addlegendentry{Random F};
    \addplot[SeaGreen,     solid, mark options={solid}, mark=*,   mark size=0.6, line width=0.7, select coords between index={0}{10}] table[x=al, y=entropy_filt_k1] \filtablacc;\addlegendentry{Entropy F};
    \addplot[Goldenrod,     solid, mark options={solid}, mark=*,   mark size=0.6, line width=0.7, select coords between index={0}{10}] table[x=al, y=coreset_filt_k1] \filtablacc;\addlegendentry{CoreSet F};
    \addplot[color=black,     dashed, mark options={solid}, mark=*,   mark size=0.6, line width=0.7, select coords between index={0}{10}] table[x=al, y=vr_nofilt_k1] \filtablacc;\addlegendentry{VR NF};
    \addplot[Magenta,     dashed, mark options={solid}, mark=*,   mark size=0.6, line width=0.7, select coords between index={0}{10}] table[x=al, y=random_nofilt_k1] \filtablacc;\addlegendentry{Random NF};
    \addplot[SeaGreen,     dashed, mark options={solid}, mark=*,   mark size=0.6, line width=0.7, select coords between index={0}{10}] table[x=al, y=entropy_nofilt_k1] \filtablacc;\addlegendentry{Entropy NF};
    \addplot[Goldenrod,     dashed, mark options={solid}, mark=*,   mark size=0.6, line width=0.7, select coords between index={0}{10}] table[x=al, y=coreset_nofilt_k1] \filtablacc;\addlegendentry{CoreSet NF};
\end{axis}
\end{tikzpicture}
\hspace{-100pt}
&
\begin{tikzpicture}
\begin{axis}[%
	width=145pt,
	height=120pt,
	xlabel={\scriptsize acquisition round $t$},
	ylabel={\scriptsize inlier rate},
	ylabel style={yshift=-5pt},
    y tick label style={xshift=2pt},
    xtick={0,5,...,10},
    xmax = 10,
    xmin = -1,
    ymax=86,ymin=0,
    legend pos= {south east},
    legend style={cells={anchor=east}, font =\scriptsize, fill opacity=0.8, row sep=-2.5pt},
    title style={align=center},
    title={\footnotesize filtering outliers \vs no filtering},
	title style={yshift=-5pt},    
	xlabel style={yshift=0pt},    
    legend style={cells={anchor=east}, font =\tiny, fill opacity=0.8, row sep=-2.5pt},
    grid=both,
    minor tick num=4,    
]
    \addplot[color=black,     dashed, mark options={solid}, mark=*,   mark size=0.6, line width=0.7, select coords between index={0}{10}] table[x=al, y=vr_nofilt_k1] \filtablid;%
    \addplot[Magenta,     dashed, mark options={solid}, mark=*,   mark size=0.6, line width=0.7, select coords between index={0}{10}] table[x=al, y=random_nofilt_k1] \filtablid;%
    \addplot[SeaGreen,     dashed, mark options={solid}, mark=*,   mark size=0.6, line width=0.7, select coords between index={0}{10}] table[x=al, y=entropy_nofilt_k1] \filtablid;%
    \addplot[Goldenrod,     dashed, mark options={solid}, mark=*,   mark size=0.6, line width=0.7, select coords between index={0}{10}] table[x=al, y=coreset_nofilt_k1] \filtablid;%
    \addplot[color=black,     solid, mark options={solid}, mark=*,   mark size=0.6, line width=0.7, select coords between index={0}{10}] table[x=al, y=vr_filt_k1] \filtablid;%
    \addplot[Magenta,     solid, mark options={solid}, mark=*,   mark size=0.6, line width=0.7, select coords between index={0}{10}] table[x=al, y=random_filt_k1] \filtablid;%
    \addplot[SeaGreen,     solid, mark options={solid}, mark=*,   mark size=0.6, line width=0.7, select coords between index={0}{10}] table[x=al, y=entropy_filt_k1] \filtablid;%
    \addplot[Goldenrod,     solid, mark options={solid}, mark=*,   mark size=0.6, line width=0.7, select coords between index={0}{10}] table[x=al, y=coreset_filt_k1] \filtablid;%
\end{axis}
\end{tikzpicture}
\hspace{1pt}
&
\begin{tikzpicture}
\begin{axis}[%
	width=145pt,
	height=120pt,
	xlabel={\scriptsize acquisition round $t$},
	ylabel={\scriptsize classification accuracy},
	ylabel style={yshift=-5pt},
    y tick label style={xshift=2pt},
    xtick={0,5,...,10},
    xmax = 11,
    xmin = -1,
    ymax = 72,
    legend columns = 4, legend pos= {south east},
    legend style={cells={anchor=west}, font =\scriptsize, fill opacity=0.8, row sep=-2.5pt, xshift=25ex, yshift=20ex},
    title style={align=center},
    title={\footnotesize $K$-way \vs $K\hspace{-3pt}+\hspace{-3pt}1$-way},
	title style={yshift=-5pt},    
	xlabel style={yshift=0pt},    
    grid=both,
    minor tick num=4,    
]
    \addplot[color=black,     solid, mark=*,   mark size=0.6, line width=0.7, select coords between index={0}{10}] table[x=al, y=vr_nofilt_k1] \filtablacc;\addlegendentry{VR K+1};
    \addplot[Magenta,     solid, mark=*,   mark size=0.6, line width=0.7, select coords between index={0}{10}] table[x=al, y=random_nofilt_k1] \filtablacc;\addlegendentry{Random K+1};
    \addplot[SeaGreen,     solid, mark=*,   mark size=0.6, line width=0.7, select coords between index={0}{10}] table[x=al, y=entropy_nofilt_k1] \filtablacc;\addlegendentry{Entropy K+1};
    \addplot[Goldenrod,     solid, mark=*,   mark size=0.6, line width=0.7, select coords between index={0}{10}] table[x=al, y=coreset_nofilt_k1] \filtablacc;\addlegendentry{CoreSet K+1};
	\addplot[color=black,     dashed, mark options={solid}, mark=*,   mark size=0.6, line width=0.7, select coords between index={0}{10}] table[x=al, y=vr_nofilt_k] \filtablacc;\addlegendentry{VR K};
    \addplot[Magenta,     dashed, mark options={solid}, mark=*,   mark size=0.6, line width=0.7, select coords between index={0}{10}] table[x=al, y=random_nofilt_k] \filtablacc;\addlegendentry{Random K};
    \addplot[SeaGreen,     dashed, mark options={solid}, mark=*,   mark size=0.6, line width=0.7, select coords between index={0}{10}] table[x=al, y=entropy_nofilt_k] \filtablacc;\addlegendentry{Entropy K};
    \addplot[Goldenrod,     dashed, mark options={solid}, mark=*,   mark size=0.6, line width=0.7, select coords between index={0}{10}] table[x=al, y=coreset_nofilt_k] \filtablacc;\addlegendentry{CoreSet K};
\end{axis}
\end{tikzpicture}
&
\hspace{-100pt}
\begin{tikzpicture}
\begin{axis}[%
	width=145pt,
	height=120pt,
	xlabel={\scriptsize acquisition round $t$},
	ylabel={\scriptsize inlier rate},
	ylabel style={yshift=-5pt},
    y tick label style={xshift=2pt},
    xtick={0,5,...,10},
    xmax = 10,
    xmin = -1,
    ymax=86,ymin=0,
    legend pos= {south east},
    legend style={cells={anchor=east}, font =\scriptsize, fill opacity=0.8, row sep=-2.5pt},
    title style={align=center},
    title={\footnotesize $K$-way \vs $K\hspace{-3pt}+\hspace{-3pt}1$-way},
	title style={yshift=-5pt},    
	xlabel style={yshift=0pt},    
    grid=both,
    minor tick num=4,    
]
    \addplot[color=black,     solid, mark=*,   mark size=0.6, line width=0.7, select coords between index={0}{10}] table[x=al, y=vr_nofilt_k1] \filtablid;%
    \addplot[Magenta,     solid, mark=*,   mark size=0.6, line width=0.7, select coords between index={0}{10}] table[x=al, y=random_nofilt_k1] \filtablid;%
    \addplot[SeaGreen,     solid, mark=*,   mark size=0.6, line width=0.7, select coords between index={0}{10}] table[x=al, y=entropy_nofilt_k1] \filtablid;%
    \addplot[Goldenrod,     solid, mark=*,   mark size=0.6, line width=0.7, select coords between index={0}{10}] table[x=al, y=coreset_nofilt_k1] \filtablid;%
	\addplot[color=black,     dashed, mark options={solid}, mark=*,   mark size=0.6, line width=0.7, select coords between index={0}{10}] table[x=al, y=vr_nofilt_k] \filtablid;%
    \addplot[Magenta,     dashed, mark options={solid}, mark=*,   mark size=0.6, line width=0.7, select coords between index={0}{10}] table[x=al, y=random_nofilt_k] \filtablid;%
    \addplot[SeaGreen,     dashed, mark options={solid}, mark=*,   mark size=0.6, line width=0.7, select coords between index={0}{10}] table[x=al, y=entropy_nofilt_k] \filtablid;%
    \addplot[Goldenrod,     dashed, mark options={solid}, mark=*,   mark size=0.6, line width=0.7, select coords between index={0}{10}] table[x=al, y=coreset_nofilt_k] \filtablid;%
\end{axis}
\end{tikzpicture}
\end{tabular}
  \vspace{-10pt} %
  \caption{Performance and inlier rate comparison for the proposed approach with different scoring functions for two different experiments. Left: with or without outlier filtering. Right: with a $K$-way or a $K+1$-way classifier during semi-supervised learning (step 3), without outlier filtering in both cases. NF: no outlier filtering. F: outlier filtering is used. Experiments on ImageNet with 0.8 outlier ratio.
  } \label{fig:filtering_ablation}
  \vspace{-5pt} %
\end{figure*}

\textbf{Comparison with other methods: }
\looseness=-1
We perform extensive experiments on ImageNet, which we consider as the most realistic setup due to the normal-sized images and the larger number of categories in the outlier class.
In Figure~\ref{fig:main_fig_acc} and Figure~\ref{fig:main_fig_idr}, we present classification accuracy and inlier rate, respectively, for  baselines, SoA approaches, and our method.
Our approach achieves the best results among all setups by a large margin. Semi-supervision is a key ingredient, as shown in the work of Simeoni \etal~\cite{oriane} too; their case was only the 0.0 outlier ratio. 
Even without semi-supervision (steps 2 and 3 in Figure~\ref{fig:overview} are skipped), our method is either on par or better than others, especially in the presence of more outliers. This happens without achieving the higher inlier rate. Choosing mostly inliers does not mean that the acquired examples are useful for training.
For instance, LfOSA achieves the highest inlier rate but performs poorly in the low presence of outliers. 
Other methods (SIMILAR, CCAL) achieve moderate inlier rates, as ours, but CCAL performs poorly, while SIMILAR is the top-performing competitor. Nevertheless, it is costly to run and not scalable (we could not run it for the 0.9 outlier ratio). MQNet performs well on low outlier ratios but fails in the large presence of outliers.
BADGE and Random perform well only without any or with few outliers, as expected. 
Overall, note that performance differences get larger for more outliers, in the challenging setups, while many methods are well-performing for the low presence of outliers.

We report the same comparison with the top competitors on CIFAR100 and TinyImageNet in Figure~\ref{fig:cifar_tiny} to confirm the good performance of the proposed method in two additional datasets. Note that both of these are with tiny resolution images, which forms the most standard setup in prior work for active learning with outliers.
CCAL performs on par with SIMILAR, presumably because using two SSL-based networks is a choice tailored to these benchmarks.

\begin{figure*}[t]
  \centering
  \vspace{-5pt}
  \input{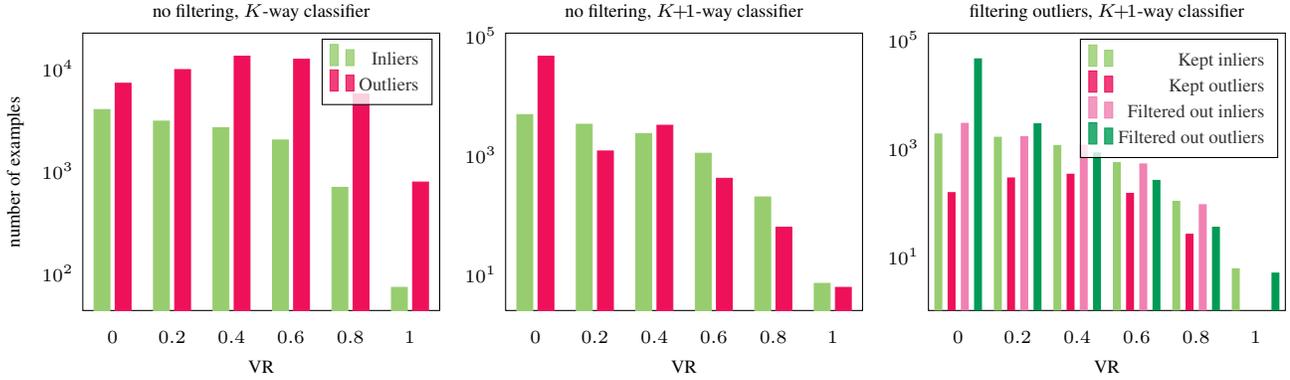}
\pgfplotsset{every tick label/.append style={font=\scriptsize}}
\pgfplotsset{select coords between index/.style 2 args={
    x filter/.code={
        \ifnum\coordindex<#1\def\pgfmathresult{}\fi
        \ifnum\coordindex>#2\def\pgfmathresult{}\fi
    }
}}
\hspace{-10pt}
\begin{tabular}{@{\msp}c@{\msp}c@{\msp}c}
\begin{tikzpicture}
\begin{axis}[%
width=180pt,
height=150pt,
ybar,
xtick style={draw=none},
ytick style={draw=none},
xtick={0,0.2,0.4,0.6,0.8,1},
ymode=log,
grid=none,
bar width=6pt,
xlabel={\scriptsize VR},%
ylabel={\scriptsize number of examples},%
legend style={cells={anchor=east}, font =\scriptsize, fill opacity=0.8, row sep=-1.5pt},
title={\scriptsize no filtering, $K$-way classifier},
title style={yshift=-5pt}
]
\addplot[color=YellowGreen,fill=YellowGreen, line width=0.0] table[x=x, y=id] \vrnofiltk;\addlegendentry{Inliers};
\addplot[color=OrangeRed,fill=OrangeRed, line width=0.0] table[x=x, y=ood] \vrnofiltk;\addlegendentry{Outliers};
\end{axis}
\end{tikzpicture}
&
\begin{tikzpicture}
\begin{axis}[%
width=180pt,
height=150pt,
ybar,
xtick style={draw=none},
ytick style={draw=none},
xtick={0,0.2,0.4,0.6,0.8,1},
ymode=log,
grid=none,
bar width=6pt,
xlabel={\scriptsize VR},%
title={\scriptsize no filtering, $K\hspace{-3pt}+\hspace{-3pt}1$-way classifier},
title style={yshift=-5pt}
]
\addplot[color=YellowGreen,fill=YellowGreen, line width=0.0] table[x=x, y=id] \vrnofiltkone;%
\addplot[color=OrangeRed,fill=OrangeRed, line width=0.0] table[x=x, y=ood] \vrnofiltkone;%
\end{axis}
\end{tikzpicture}
&
\begin{tikzpicture}
\begin{axis}[%
width=180pt,
height=150pt,
ybar,
xtick style={draw=none},
ytick style={draw=none},
xtick={0,0.2,0.4,0.6,0.8,1},
ymin = 1,
ymode=log,
grid=none,
bar width=3pt,
xlabel={\scriptsize VR},%
legend style={cells={anchor=east}, font =\scriptsize, fill opacity=0.8, row sep=-1.5pt},
title={\scriptsize filtering outliers, $K\hspace{-3pt}+\hspace{-3pt}1$-way classifier},
title style={yshift=-5pt}
]
\addplot[color=YellowGreen,fill=YellowGreen, draw=none] table[x=x, y=kept_id] \vrfiltkone;\addlegendentry{Kept inliers};
\addplot[color=OrangeRed,fill=OrangeRed, draw=none] table[x=x, y=kept_ood] \vrfiltkone;\addlegendentry{Kept outliers};
\addplot[color=CarnationPink,fill=CarnationPink, draw=none] table[x=x, y=filtered_id] \vrfiltkone;\addlegendentry{Filtered out inliers};
\addplot[color=ForestGreen,fill=ForestGreen, draw=none] table[x=x, y=filtered_ood] \vrfiltkone;\addlegendentry{Filtered out outliers};
\end{axis}
\end{tikzpicture}
\end{tabular}
  \vspace{-10pt} %
  \caption{Histograms of VR values for unlabeled examples with our method during round 3 and an outlier ratio of 0.8 on ImageNet. Histograms are created separately for inliers and outliers, and for filtered out or kept inliers/outliers in the case outlier filtering is used. 
  } \label{fig:histograms}
  \vspace{-0pt} %
\end{figure*}

\begin{figure}[t]
  \centering
  \vspace{-5pt}
  \input{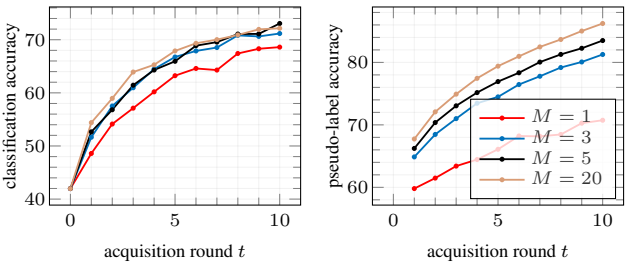}
\pgfplotsset{every tick label/.append style={font=\scriptsize}}
\pgfplotsset{select coords between index/.style 2 args={
    x filter/.code={
        \ifnum\coordindex<#1\def\pgfmathresult{}\fi
        \ifnum\coordindex>#2\def\pgfmathresult{}\fi
    }
}}
\pgfplotsset{minor grid style={solid,gray,opacity=0.1}}
\pgfplotsset{major grid style={solid,gray,opacity=0.1}}
\begin{tabular}{@{\nssp}c@{\msp}c@{\nssp}}
\begin{tikzpicture}
\begin{axis}[%
	width=140pt,
	height=120pt,
	xlabel={\scriptsize acquisition round $t$},
	ylabel={\scriptsize classification accuracy},
	ylabel style={yshift=-5pt},
    y tick label style={xshift=2pt},
    xtick={0,5,...,10},
    xmax = 11,
    xmin = -1,
    legend pos= {south east},
    legend style={cells={anchor=west}, font =\scriptsize, fill opacity=0.8, row sep=-2.5pt},
    title style={align=center},
	title style={yshift=-5pt},    
	xlabel style={yshift=0pt},    
    grid=both,
    minor tick num=4,    
]
    \addplot[color=red,     solid, mark=*,   mark size=0.6, line width=0.7, select coords between index={0}{10}] table[x=al, y=ens1] \ensabl;%
	\addplot[color=RoyalBlue,     solid, mark=*,   mark size=0.6, line width=0.7, select coords between index={0}{10}] table[x=al, y=ens3] \ensabl;%
	\addplot[color=black,     solid, mark=*,   mark size=0.6, line width=0.7, select coords between index={0}{10}] table[x=al, y=ens5] \ensabl;%
	\addplot[color=Tan,     solid, mark=*,   mark size=0.6, line width=0.7, select coords between index={0}{10}] table[x=al, y=ens20] \ensabl;%
\end{axis}
\end{tikzpicture}
&
\begin{tikzpicture}
\begin{axis}[%
	width=140pt,
	height=120pt,
	xlabel={\scriptsize acquisition round $t$},
	ylabel={\scriptsize pseudo-label accuracy},
	ylabel style={yshift=-5pt},
    y tick label style={xshift=2pt},
    xtick={0,5,...,10},
    xmax = 11,
    xmin = -1,
    legend pos= {south east},
    legend style={cells={anchor=west}, font =\scriptsize, fill opacity=0.8, row sep=-2.5pt},
    title style={align=center},
	title style={yshift=-5pt},    
	xlabel style={yshift=0pt},    
    grid=both,
    minor tick num=4,    
]
    \addplot[color=red,     solid, mark=*,   mark size=0.6, line width=0.7, select coords between index={0}{10}] table[x=al, y=ens1] \plabacc;\addlegendentry{$M = 1$};
	\addplot[color=RoyalBlue,     solid, mark=*,   mark size=0.6, line width=0.7, select coords between index={0}{10}] table[x=al, y=ens3] \plabacc;\addlegendentry{$M = 3$};
	\addplot[color=black,     solid, mark=*,   mark size=0.6, line width=0.7, select coords between index={0}{10}] table[x=al, y=ens5] \plabacc;\addlegendentry{$M = 5$};/
	\addplot[color=Tan,     solid, mark=*,   mark size=0.6, line width=0.7, select coords between index={0}{10}] table[x=al, y=ens20] \plabacc;\addlegendentry{$M = 20$};
\end{axis}
\end{tikzpicture}
\end{tabular}
  \vspace{-5pt} %
  \caption{Classification accuracy (left) and accuracy of the predicted pseudo-labels (right) for increasing the size of the ensemble on ImageNet for 0.5 outlier ratio. The scoring function is VR for all cases with $M>1$ and random for $M=1$, where VR is not defined.
   \label{fig:ensemble}}
  \vspace{-0pt} %
\end{figure}

\textbf{Impact of training with outliers: }
First, we validate the importance of outlier filtering for different acquisition functions. To perform that, we perform acquisition with and without outlier filtering, \ie setting predicted outliers to zero score or not. Results are presented in Figure~\ref{fig:filtering_ablation} (left).
Simple scoring functions such as Random and Entropy benefit a lot from outlier filtering, which increases their performance and inlier rate by a large margin.
This is much less the case for CoreSet, and nearly not at all for VR. 
Moreover, with filtering, even simple Entropy is top performing.

We go one step further and investigate performance without filtering for the standard choice of jointly training a $K+1$-way classifier and compare it with the case of using a $K$-way classifier for the acquisition part. That means that during step 3 in semi-supervised learning, we only train with labeled and pseudo-labeled inliers. 
Results are presented in Figure~\ref{fig:filtering_ablation} on the right. It turns out that there is a large performance drop if outliers are not used in the training. It is the joint training that makes the outlier filtering less necessary. 

To take a closer look, we present an analysis of the VR values in Figure~\ref{fig:histograms}. Without filtering, VR with $K$-way acquires a large number of outliers (left), while it acquires noticeably fewer outliers with the joint $K+1$-way training despite no filtering (middle). 
We explain this by the fact that inlier examples from different classes are more similar to each other than to outliers examples. Therefore, inlier examples more often result in disagreements. This may be seen as an outcome of the chosen inlier classes, which nevertheless follows prior work~\cite{ccal} and imitates a realistic scenario.
The outlier filtering (right) does not significantly improve acquisition because it operates better on the low VR regime, where more outliers than inliers are rejected. In the high VR regime, where acquired examples belong, an equal amount of inliers is filtered out too.

\textbf{Impact of ensembles: }
In Figure~\ref{fig:ensemble}, we show the impact of the ensemble size on classification accuracy.  
Using a network ensemble has a significant impact on performance already from $M=3$. The increased pseudo-label accuracy is the main source of the improvement.
Training 20 networks have good benefits in early rounds. Nevertheless, we opt for $M=5$ as the standard setup to lower training complexity and experimentation time. Let us note once more that ensembles are not used for evaluating test accuracy, where a single network is used.

\begin{figure}[t]
  \centering
  \vspace{-5pt} %
  \input{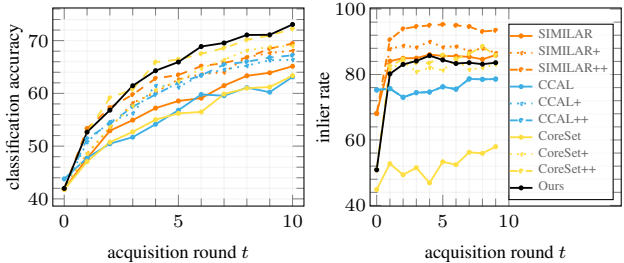}
\pgfplotsset{every tick label/.append style={font=\scriptsize}}
\pgfplotsset{select coords between index/.style 2 args={
    x filter/.code={
        \ifnum\coordindex<#1\def\pgfmathresult{}\fi
        \ifnum\coordindex>#2\def\pgfmathresult{}\fi
    }
}}
\pgfplotsset{minor grid style={solid,gray,opacity=0.1}}
\pgfplotsset{major grid style={solid,gray,opacity=0.1}}
\pgfplotsset{
compat=1.11,
legend image code/.code={
\draw[mark repeat=2,mark phase=2]
plot coordinates {
(0cm,0cm)
(0.15cm,0cm)        %
(0.3cm,0cm)         %
};%
}
}
\hspace{-18pt}
\begin{tabular}{@{\msp}c@{\msp}c}
\begin{tikzpicture}
\begin{axis}[%
	width=140pt,
	height=120pt,
	xlabel={\scriptsize acquisition round $t$},
	ylabel={\scriptsize classification accuracy},
	ylabel style={yshift=-5pt},
    y tick label style={xshift=2pt},
    xtick={0,5,...,10},
    xmax = 10.5,
    xmin = -0.5,
    legend pos= {south east},
    legend style={cells={anchor=west}, font =\scriptsize, fill opacity=0.6, row sep=-2.5pt},
    title style={align=center, font=\scriptsize, row sep=3pt},
	title style={yshift=-5pt},    
	xlabel style={yshift=0pt},    
    grid=both,
    minor tick num=4,
]
	\addplot[color=orange,solid, mark=*,   mark size=0.6, line width=0.7, select coords between index={0}{10}] table[x=al, y=original_similar] \stepstoother;%
    \addplot[color=orange,dotted, mark=*,   mark size=0.6, line width=0.7, select coords between index={0}{10}] table[x=al, y=ssl_similar] \stepstoother;%
    \addplot[color=orange,dashed, mark=*,   mark size=0.6, line width=0.7, select coords between index={0}{10}] table[x=al, y=ens_similar] \stepstoother;%
	\addplot[color=CornflowerBlue,solid, mark=*,   mark size=0.6, line width=0.7, select coords between index={0}{10}] table[x=al, y=original_ccal] \stepstoother;%
    \addplot[color=CornflowerBlue,dotted, mark=*,   mark size=0.6, line width=0.7, select coords between index={0}{10}] table[x=al, y=ssl_ccal] \stepstoother;%
    \addplot[color=CornflowerBlue,dashed, mark=*,   mark size=0.6, line width=0.7, select coords between index={0}{10}] table[x=al, y=ens_ccal] \stepstoother;%
    \addplot[Goldenrod,     solid, mark=*,   mark size=0.6, line width=0.7, select coords between index={0}{10}] table[x=al, y=original_coreset] \stepstoother;%
    \addplot[Goldenrod,     dotted, mark=*,   mark size=0.6, line width=0.7, select coords between index={0}{10}] table[x=al, y=ssl_coreset] \stepstoother;%
    \addplot[Goldenrod,     dashed, mark=*,   mark size=0.6, line width=0.7, select coords between index={0}{10}] table[x=al, y=ens_coreset] \stepstoother;%
	\addplot[color=black,solid, mark=*,   mark size=0.6, line width=0.7, select coords between index={0}{10}] table[x=al, y=ours_semi] \stepstoother;%
\end{axis}
\end{tikzpicture}
&
\hspace{-10pt}
\begin{tikzpicture}
\begin{axis}[%
	width=140pt,
	height=120pt,
	xlabel={\scriptsize acquisition round $t$},
	ylabel={\scriptsize inlier rate},
	ylabel style={yshift=-5pt},
    y tick label style={xshift=2pt},
    xtick={0,5,...,10},
    xmax = 18.5,
    xmin = -0.5,
    legend pos= {south east},
    legend style={cells={anchor=west}, font =\tiny, fill opacity=0.6, row sep=-2.5pt, inner sep=0pt},
    title style={align=center, font=\scriptsize, row sep=3pt},
	title style={yshift=-5pt},    
	xlabel style={yshift=0pt},    
    grid=both,
    minor tick num=4,
]
	\addplot[color=orange,solid, mark=*,   mark size=0.6, line width=0.7, select coords between index={0}{10}] table[x=al, y=original_similar] \stepstootherid;\addlegendentry{SIMILAR};
    \addplot[color=orange,dotted, mark=*,   mark size=0.6, line width=0.7, select coords between index={0}{10}] table[x=al, y=ssl_similar] \stepstootherid;\addlegendentry{SIMILAR+};
    \addplot[color=orange,dashed, mark=*,   mark size=0.6, line width=0.7, select coords between index={0}{10}] table[x=al, y=ens_similar] \stepstootherid;\addlegendentry{SIMILAR++};
	\addplot[color=CornflowerBlue,solid, mark=*,   mark size=0.6, line width=0.7, select coords between index={0}{10}] table[x=al, y=original_ccal] \stepstootherid;\addlegendentry{CCAL};
    \addplot[color=CornflowerBlue,dotted, mark=*,   mark size=0.6, line width=0.7, select coords between index={0}{10}] table[x=al, y=ssl_ccal] \stepstootherid;\addlegendentry{CCAL+};
    \addplot[color=CornflowerBlue,dashed, mark=*,   mark size=0.6, line width=0.7, select coords between index={0}{10}] table[x=al, y=ens_ccal] \stepstootherid;\addlegendentry{CCAL++};
    \addplot[Goldenrod,     solid, mark=*,   mark size=0.6, line width=0.7, select coords between index={0}{10}] table[x=al, y=original_coreset] \stepstootherid;\addlegendentry{CoreSet};
    \addplot[Goldenrod,     dotted, mark=*,   mark size=0.6, line width=0.7, select coords between index={0}{10}] table[x=al, y=ssl_coreset] \stepstootherid;\addlegendentry{CoreSet+};
    \addplot[Goldenrod,     dashed, mark=*,   mark size=0.6, line width=0.7, select coords between index={0}{10}] table[x=al, y=ens_coreset] \stepstootherid;\addlegendentry{CoreSet++};
	\addplot[color=black,solid, mark=*,   mark size=0.6, line width=0.7, select coords between index={0}{10}] table[x=al, y=ours_semi] \stepstootherid;\addlegendentry{Ours};
\end{axis}
\end{tikzpicture}
\end{tabular}
  \vspace{-5pt} %
  \caption{
  Impact of adding our key ingredients to other methods. Performance and inlier rate comparison when joint training and semi-supervision (denoted by +) and ensembles (all three ingredients together, denoted by ++) are combined with other methods. Experiments on ImageNet with 0.5 outlier ratio.
  \label{fig:steps_others}
  }
  \vspace{-0pt} %
\end{figure}

\begin{figure}[t]
  \centering
  \input{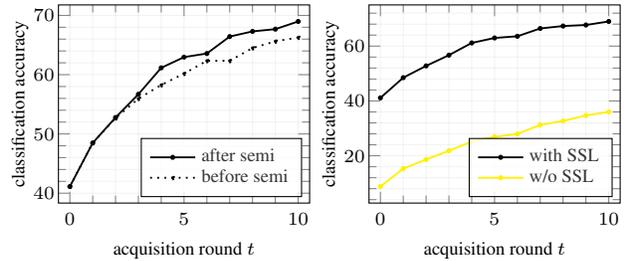}
\pgfplotsset{every tick label/.append style={font=\scriptsize}}
\pgfplotsset{select coords between index/.style 2 args={
    x filter/.code={
        \ifnum\coordindex<#1\def\pgfmathresult{}\fi
        \ifnum\coordindex>#2\def\pgfmathresult{}\fi
    }
}}
\pgfplotsset{minor grid style={solid,gray,opacity=0.1}}
\pgfplotsset{major grid style={solid,gray,opacity=0.1}}
\hspace{-18pt}
\begin{tabular}{@{\msp}c@{\msp}c}
\begin{tikzpicture}
\begin{axis}[%
    width=140pt,
	height=120pt,
	xlabel={\scriptsize acquisition round $t$},
	ylabel={\scriptsize classification accuracy},
	ylabel style={yshift=-5pt},
    y tick label style={xshift=2pt},
    xtick={0,5,...,10},
    xmax = 10.5,
    xmin = -0.5,
    legend pos= {south east},
    legend style={cells={anchor=west}, font =\scriptsize, fill opacity=0.8, row sep=-2.5pt},
    title style={align=center, font=\scriptsize, row sep=3pt},
	title style={yshift=-5pt},    
	xlabel style={yshift=0pt},    
    grid=both,
    minor tick num=4,
]
	\addplot[color=black,solid, mark=*,   mark size=0.6, line width=0.7, select coords between index={0}{10}] table[x=al, y=ours_80] \acqbeforessl;\addlegendentry{after semi};
    \addplot[color=black,dotted, mark=*,   mark size=0.6, line width=0.7, select coords between index={0}{10}] table[x=al, y=ours_acq_ssl_80] \acqbeforessl;\addlegendentry{before semi};
\end{axis}
\end{tikzpicture}
&
\hspace{-10pt}
\begin{tikzpicture}
\begin{axis}[%
	width=140pt,
	height=120pt,
	xlabel={\scriptsize acquisition round $t$},
	ylabel={\scriptsize classification accuracy},
	ylabel style={yshift=-5pt},
    y tick label style={xshift=2pt},
    xtick={0,5,...,10},
    xmax = 10.5,
    xmin = -0.5,
    legend pos= {south east},
    legend style={cells={anchor=west}, font =\scriptsize, fill opacity=0.8, row sep=-2.5pt},
    title style={align=center, font=\scriptsize, row sep=3pt},
	title style={yshift=-5pt},    
	xlabel style={yshift=0pt},    
    grid=both,
    minor tick num=4,
]
	\addplot[color=black,     solid, mark=*,   mark size=0.6, line width=0.7, select coords between index={0}{10}] table[x=al, y=ours_semi] \ineightyood;\addlegendentry{with SSL};
    \addplot[color=yellow,     solid, mark=*,   mark size=0.6, line width=0.7, select coords between index={0}{10}] table[x=al, y=no_ssl] \nossl;\addlegendentry{w/o SSL};
\end{axis}
\end{tikzpicture}
\end{tabular}
  \vspace{-5pt} %
  \caption{
  Left: classification accuracy when the acquisition is performed before and after (default) training with semi-supervision. Right: impact of self-supervised pre-training on classification performance. Experiments on ImageNet with 0.8 outlier ratio.
  \label{fig:acq_ablation}
  }
  \vspace{0pt} %
\end{figure}

\textbf{Our components improve other methods:}
We investigate whether our key components, \ie joint training, ensembles, and semi-supervision, are beneficial to other methods too. In particular, we implement the combination for CCAL, SIMILAR, and CoreSet. 
We add an outlier class to the classifier for CoreSet and CCAL, while SIMILAR already includes it.
This allows us to perform pseudo-labeling and semi-supervision.
For CoreSet and SIMILAR to benefit from model ensembling, we use average features and similarity matrices, respectively.
CCAL performs acquisition based on a fixed network obtained during pre-training; therefore, semi-supervision or ensembling do not affect its acquisition phase.
Results in Figure~\ref{fig:steps_others} show that all methods benefit from our key ingredients.
However, compared to much more complicated methods, our simple and intuitive approach is top-performing. 
Note that the top inlier rate does not necessarily result in top performance.
Additionally, we see that classical active learning methods (CoreSet) become competitive again.

\vspace{15pt}

\textbf{Impact of semi-supervision on the acquisition:}
In Figure~\ref{fig:acq_ablation} (left), we present the impact of performing acquisition before or after (default) semi-supervision. In the case of performing acquisition before semi-supervision, we still use semi-supervision to train the network used for evaluation. Results show that semi-supervision improves acquisition, although the difference is small in the early rounds.

\vspace{10pt}

\textbf{Impact of self-supervised pre-training: }
We present the impact of self-supervised pre-training on classification accuracy in Figure~\ref{fig:acq_ablation} (right). Results show that it provides a significant benefit of 30\% in round 0. This performance gap is retained across all active learning rounds too. This result confirms the findings of earlier work~\cite{oriane}.

\section{Conclusions}
\label{sec:conclusions}
We improve the state-of-the-art performance on active learning with outliers by a large margin.
This is achieved by three ingredients: joint training with inliers and outliers, semi-supervision via pseudo-labeling, and network ensembles that are used in a way not to increase the test-time complexity.
Some of our findings are shown to be compatible with existing acquisition functions, and their applicability goes beyond existing approaches due to the universality of the proposed framework. 
Simple acquisition functions that were thought to fail in this setup are able to reach state-of-the-art performance within our framework. 
We will publicly release the source code and datasets of our extensive evaluation for reproducibility and to improve the setup discrepancy from which the current literature suffers.

{
\small
\bibliographystyle{ieee_fullname}
\bibliography{tex/bib}
}

\appendix

\section{Results on the setup from MQNet }
\looseness=-1
We evaluate our method on the experimental setup of MQNet~\cite{mqnet} and present the results in Table~\ref{tab:mqnet_data}. 
We perform this experiment due to the following differences: (1) to use the same inlier/outlier class splits and the same initial labels set and unlabeled set, (2) to perform experiments without SSL pre-training as in their work and (3) to perform the network training with the same hyper-parameters as in their work. To be sure for a direct comparison, we implement our method in their own implementation framework. 
Results confirm the same observations as in our own setup; our method  outperforms MQNet even without semi-supervision.

\begin{table}[t]
    \begin{tabular}{cccc}
        \hline
        \multirow{2}{*}{Method} & \multicolumn{3}{c}{Dataset}   \\
                            & CIFAR10 & CIFAR100 & ImageNet \\
        \hline
    MQNet                   &    89.51     &   52.82       &   54.11       \\
    Ours w/o semi & 91.63 & 54.23 & 58.00 \\
         Ours & 92.95 & 56.20 & 62.30 \\
    \hline
    \end{tabular}

    \caption{Results after 10 acquisition rounds on the setup from the MQNet paper. Outlier ratio is 0.6. Reported values for MQNet are taken from~\cite{mqnet}. \label{tab:mqnet_data}}
\end{table}
\section{Impact of pseudo-label weights}
\begin{figure}[b]
  \centering
  \input{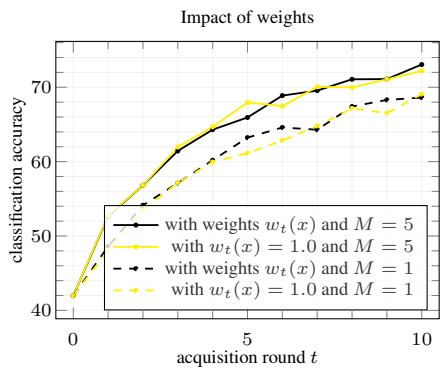}
\pgfplotsset{every tick label/.append style={font=\scriptsize}}
\pgfplotsset{select coords between index/.style 2 args={
    x filter/.code={
        \ifnum\coordindex<#1\def\pgfmathresult{}\fi
        \ifnum\coordindex>#2\def\pgfmathresult{}\fi
    }
}}
\pgfplotsset{minor grid style={solid,gray,opacity=0.1}}
\pgfplotsset{major grid style={solid,gray,opacity=0.1}}
\begin{tikzpicture}
\begin{axis}[%
	width=190pt,
	height=150pt,
	xlabel={\scriptsize acquisition round $t$},
	ylabel={\scriptsize classification accuracy},
	ylabel style={yshift=-5pt},
    y tick label style={xshift=2pt},
    x tick label style={yshift=-2pt},
    xtick={0,5,...,10},
    xmax = 10.5,
    xmin = -0.5,
    legend pos= {south east},
    legend style={cells={anchor=east}, font =\scriptsize, fill opacity=0.8, row sep=-2.5pt},
    title style={align=center, font=\scriptsize, row sep=3pt},
        title={\scriptsize{Impact of weights}},
	title style={yshift=-4pt},    
	xlabel style={yshift=5pt},    
    grid=both,
    minor tick num=4,
]
	\addplot[color=black,     solid, mark=*,   mark size=0.6, line width=0.7, select coords between index={0}{10}] table[x=al, y=ours_semi] \infiftyood;\addlegendentry{with weights $w_t(x)$ and $M=5$};
    \addplot[color=yellow,     solid, mark=*,   mark size=0.6, line width=0.7, select coords between index={0}{10}] table[x=al, y=ens5] \noweight;\addlegendentry{with $w_t(x) = 1.0$ and $M=5$};
    \addplot[color=black,     dashed, mark=*,   mark size=0.6, line width=0.7, select coords between index={0}{10}] table[x=al, y=ens1] \ensabl;\addlegendentry{with weights $w_t(x)$ and $M=1$};
    \addplot[color=yellow,     dashed, mark=*,   mark size=0.6, line width=0.7, select coords between index={0}{10}] table[x=al, y=ens1] \noweight;\addlegendentry{with $w_t(x) = 1.0$ and $M=1$};
\end{axis}
\end{tikzpicture}
  \caption{Comparison of classification accuracy for our approach with semi-supervision with and without weights for pseudo-labels in Equation~\ref{equ:loss_semi} of the main paper. Results are presented on ImageNet dataset with a 0.5 outlier ratio.
  \label{fig:ablation_w}
  }
\end{figure}

We evaluate the impact of weights $w_t(x)$ used for pseudo-labels by setting them all to $1.0$. Results are presented in Figure~\ref{fig:ablation_w}, which shows that weights provide a benefit if an ensemble is not used, while results with and without weights are comparable in the case ensemble is used. This is because ensembles improve pseudo-label accuracy, so assigning high weights for them is safe.

\begin{figure*}[t]
  \centering
  \definecolor{C1}{RGB}{141, 211, 199}
\definecolor{C2}{RGB}{255, 255, 179}
\definecolor{C3}{RGB}{190, 186, 218}
\definecolor{C4}{RGB}{128, 177, 211}
\definecolor{C5}{RGB}{128, 255, 255}

\pgfplotsset{
  bar cycle list/.style={
    cycle list={%
      {Green!50!black,fill=Green,mark=none},%
      {LimeGreen!50!black,fill=LimeGreen,mark=none},%
      {Red!50!black,fill=Red,mark=none},%
      {OrangeRed!50!black,fill=OrangeRed,mark=none},%
      {Bittersweet!50!black,fill=Bittersweet,mark=none},%
    }
  },
}
\pgfplotsset{every tick label/.append style={font=\scriptsize}}
\pgfplotsset{minor grid style={solid,gray,opacity=0.1}}
\pgfplotsset{major grid style={solid,gray,opacity=0.1}}

\centering
\small

\begin{tikzpicture}
\begin{axis}[
    ybar,
    title = {acquisition round $t= 1$},
    title style={yshift=-1.5ex},          
    enlarge y limits=upper,
    width=1.0\linewidth,
    height=0.22\linewidth,
    legend style={at={(0.5,-0.25)},anchor=north,legend columns=-1},
    ylabel={percentage},
    xlabel = {$w_t(x)$},
    symbolic x coords={0.05, 0.15, 0.25, 0.35, 0.45, 0.55, 0.65, 0.75, 0.85, 0.95},
    xtick=data,
    /pgf/bar width=5pt,%
    xtick style={draw=none},
    x tick label style={inner sep=-2pt},    
    ymajorgrids=true,
    yminorgrids=true,
    minor tick num=4,
    ymax=18,
    ]
\addplot coordinates {(0.05, 0.20) (0.15, 2.67) (0.25, 5.34) (0.35, 7.32) (0.45, 9.33) (0.55, 10.55) (0.65, 11.66) (0.75, 12.62) (0.85, 13.42) (0.95, 12.28)};
\addplot coordinates {(0.05, 0.15) (0.15, 2.83) (0.25, 6.79) (0.35, 8.58) (0.45, 8.15) (0.55, 6.81) (0.65, 5.40) (0.75, 3.56) (0.85, 2.16) (0.95, 2.71)};
\addplot coordinates {(0.05, 0.34) (0.15, 2.72) (0.25, 3.84) (0.35, 3.34) (0.45, 2.14) (0.55, 1.16) (0.65, 0.68) (0.75, 0.25) (0.85, 0.11) (0.95, 0.02)};
\addplot coordinates {(0.05, 0.40) (0.15, 3.79) (0.25, 5.47) (0.35, 4.12) (0.45, 2.30) (0.55, 1.22) (0.65, 0.41) (0.75, 0.26) (0.85, 0.06) (0.95, 0)};
\addplot coordinates {(0.05, 0.55) (0.15, 6.73) (0.25, 9.78) (0.35, 8.10) (0.45, 5.38) (0.55, 2.38) (0.65, 1.39) (0.75, 0.38) (0.85, 0.12) (0.95, 0.02)};
\end{axis}
\end{tikzpicture}

\vspace{5pt}

\begin{tikzpicture}
\begin{axis}[
    ybar,
    title = {acquisition round $t= 5$},
    title style={yshift=-1.5ex},          
    enlarge y limits=upper,
    width=1.0\linewidth,
    height=0.22\linewidth,
    legend style={at={(0.5,-0.25)},anchor=north,legend columns=-1},
    ylabel={percentage},
    xlabel = {$w_t(x)$},
    symbolic x coords={0.05, 0.15, 0.25, 0.35, 0.45, 0.55, 0.65, 0.75, 0.85, 0.95},
    xtick=data,
    /pgf/bar width=5pt,%
    xtick style={draw=none},
    x tick label style={inner sep=-2pt},    
    ymajorgrids=true,
    yminorgrids=true,
    minor tick num=4,
    ymax=18,
    ]
\addplot coordinates {(0.05, 0.04) (0.15, 1.51) (0.25, 6.07) (0.35, 11.27) (0.45, 13.69) (0.55, 13.59) (0.65, 13.08) (0.75, 11.67) (0.85, 10.88) (0.95, 9.37)};
\addplot coordinates {(0.05, 0) (0.15, 0.24) (0.25, 2.42) (0.35, 7.19) (0.45, 11.48) (0.55, 11.24) (0.65, 9.85) (0.75, 7.24) (0.85, 4.42) (0.95, 5.27)};
\addplot coordinates {(0.05, 0) (0.15, 0.20) (0.25, 1.26) (0.35, 2.47) (0.45, 2.22) (0.55, 1.35) (0.65, 0.70) (0.75, 0.44) (0.85, 0.12) (0.95, 0.08)};
\addplot coordinates {(0.05, 0.07) (0.15, 1.36) (0.25, 5.87) (0.35, 7.67) (0.45, 5.02) (0.55, 2.48) (0.65, 0.97) (0.75, 0.23) (0.85, 0.07) (0.95, 0.01)};
\addplot coordinates {(0.05, 0.01) (0.15, 0.34) (0.25, 2.74) (0.35, 4.71) (0.45, 4.38) (0.55, 2.64) (0.65, 1.36) (0.75, 0.60) (0.85, 0.09) (0.95, 0.09)};
\end{axis}
\end{tikzpicture}

\vspace{5pt}

\begin{tikzpicture}
\begin{axis}[
    ybar,
    title = {acquisition round $t= 10$},
    title style={yshift=-1.5ex},          
    enlarge y limits=upper,
    width=1.0\linewidth,
    height=0.22\linewidth,
    legend style={at={(0.5,-0.35)},anchor=north,legend columns=-1},
    ylabel={percentage},
    xlabel = {$w_t(x)$},
    symbolic x coords={0.05, 0.15, 0.25, 0.35, 0.45, 0.55, 0.65, 0.75, 0.85, 0.95},
    xtick=data,
    /pgf/bar width=5pt,%
    xtick style={draw=none},
    x tick label style={inner sep=-2pt},    
    ymajorgrids=true,
    yminorgrids=true,
    minor tick num=4,
    ymax=18,
    ]
\addplot coordinates {(0.05, 0.10) (0.15, 1.55) (0.25, 6.00) (0.35, 12.20) (0.45, 14.89) (0.55, 14.83) (0.65, 14.26) (0.75, 12.30) (0.85, 9.58) (0.95, 8.00)};
\addplot coordinates {(0.05, 0) (0.15, 0.18) (0.25, 1.54) (0.35, 5.35) (0.45, 10.35) (0.55, 12.96) (0.65, 13.48) (0.75, 11.07) (0.85, 6.50) (0.95, 8.36)};
\addplot coordinates {(0.05, 0) (0.15, 0.17) (0.25, 0.88) (0.35, 1.49) (0.45, 1.62) (0.55, 0.88) (0.65, 0.73) (0.75, 0.28) (0.85, 0.10) (0.95, 0.14)};
\addplot coordinates {(0.05, 0.02) (0.15, 0.88) (0.25, 4.23) (0.35, 6.57) (0.45, 4.53) (0.55, 2.30) (0.65, 0.72) (0.75, 0.26) (0.85, 0.09) (0.95, 0.01)};
\addplot coordinates {(0.05, 0) (0.15, 0.25) (0.25, 1.28) (0.35, 2.44) (0.45, 2.56) (0.55, 1.72) (0.65, 1.32) (0.75, 0.84) (0.85, 0.15) (0.95, 0.04)};
\legend{Correct outlier, Correct inlier, Incorrect outlier, Incorrect inlier as outlier, Incorrect inlier as inlier}
\end{axis}
\end{tikzpicture}
  \caption{Distribution of weights $w_t(x)$ for different types of pseudo-labels. \emph{Correct inlier/outlier}: example pseudo-labeled correctly. \emph{Incorrect outlier}: outlier example wrongly pseudo-labeled as an inlier (as any of the inlier classes). \emph{Incorrect inlier as outlier}: inlier example incorrectly pseudo-labeled as an outlier. \emph{Incorrect inlier as inlier}: inlier example wrongly pseudo-labeled into the wrong inlier class. Y-axis shows the percentage of outlier/inlier examples from each type, \ie \emph{Correct outlier} and \emph{Incorrect outlier} sum to 100, and \emph{Correct inlier}, \emph{Incorrect inlier as outlier} and \emph{Incorrect inlier as inlier} also sum to 100. 
  \label{fig:weights_normalized}
  }
  \vspace{-10pt} %
\end{figure*}
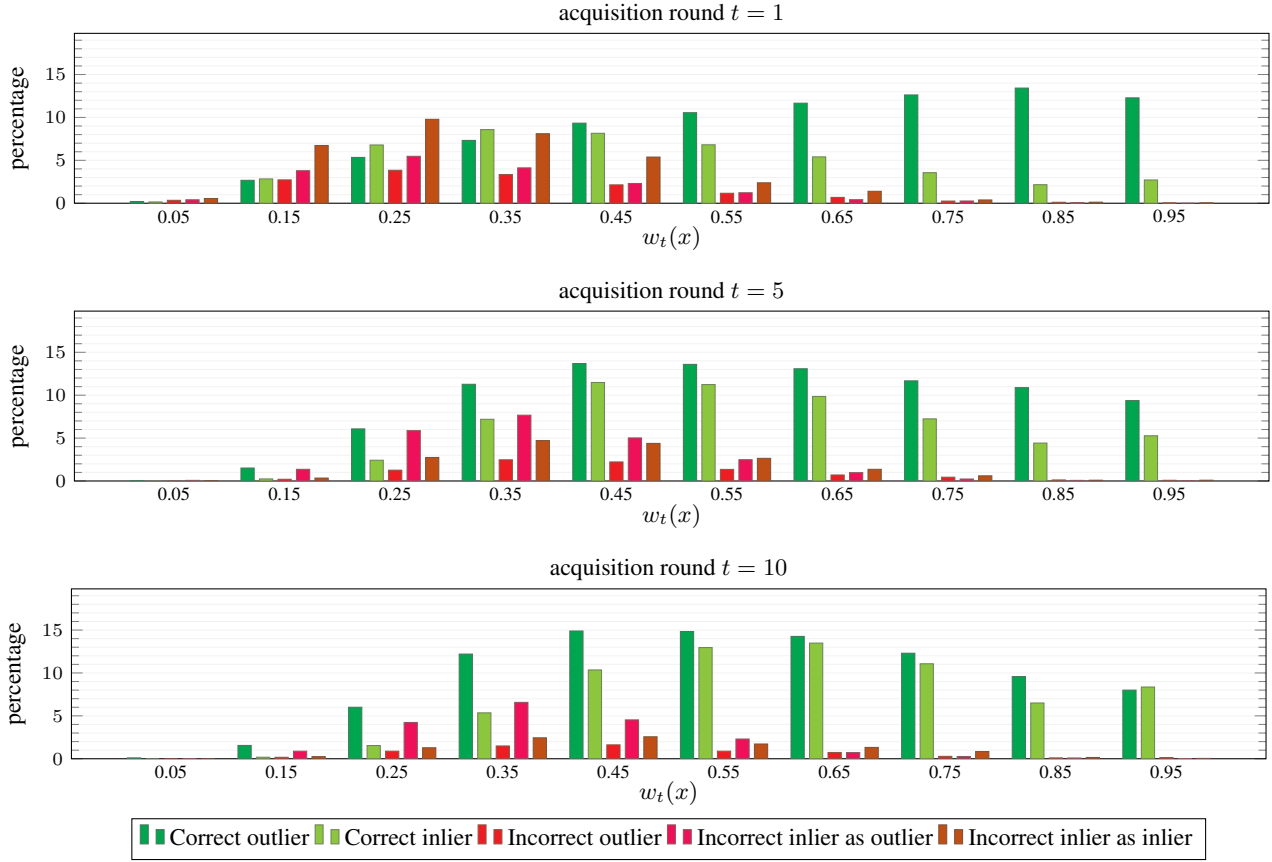

\looseness=-1
In Figure~\ref{fig:weights_normalized}, we show the evolution of pseudo-label weights over active learning rounds. It is observed that over active learning rounds, correct pseudo-labels are getting higher weights meaning that the classifier is becoming certain about those predictions. In contrast, incorrect pseudo-labels mostly have weights in the lower middle of the range.

\section{Experiment with a smaller labeled set}

\begin{figure*}[t]
  \centering
  \input{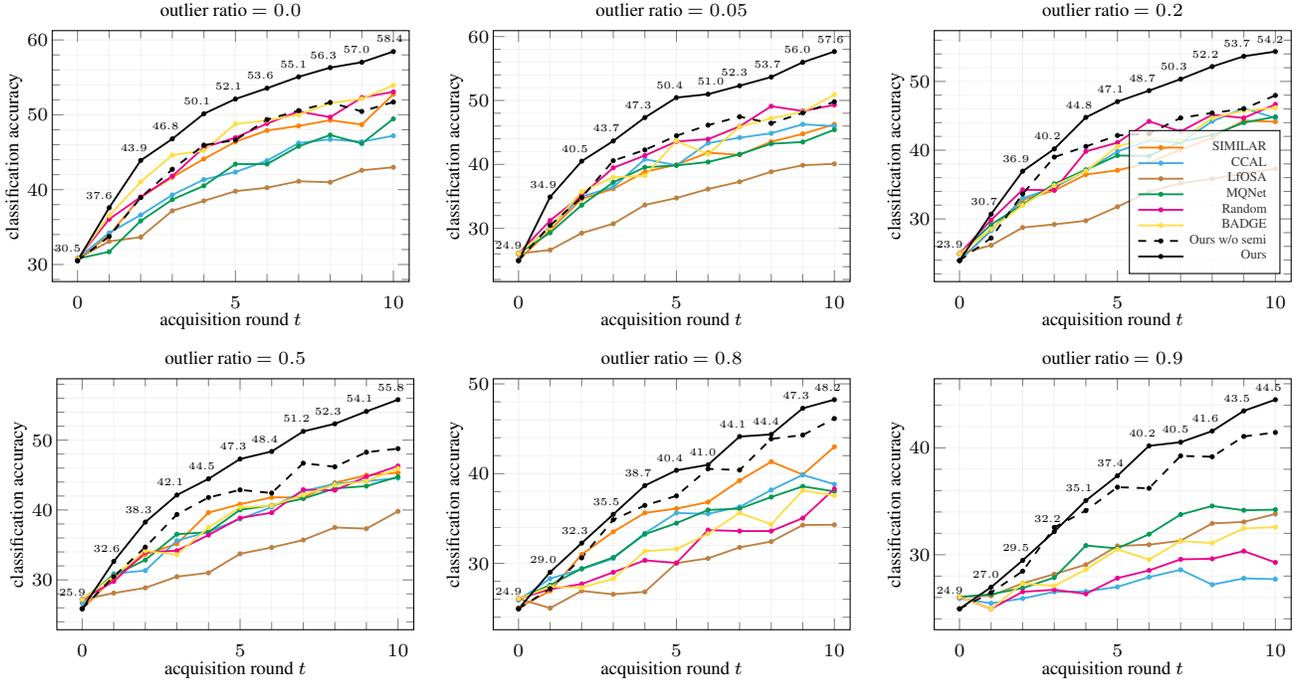}
  \vspace{-10pt} %
  \caption{Comparison of classification accuracy over multiple active learning rounds for varying outlier ratios on ImageNet when initial labeled set $L_0$ contains 5 examples (in contrast to 20 in the main paper) per inlier class and budget is equal to 100. SIMILAR is excluded for 0.9 outlier ratio since we were not able to run it even on a machine with 800GB of RAM. 
  \label{fig:five_shot}
  }
  \vspace{-10pt} %
\end{figure*}

In Figure~\ref{fig:five_shot}, we present additional results on the ImageNet dataset for the case when the initial labeled set $L_0$ contains $5$ examples per class and the budget is set to $100$. 
Results show that our approach outperforms all other recent state-of-the-art competitors and baseline methods by a large margin.  
The variant without semi-supervision is either second best or close to second best over all cases.

\section{Detailed results}

We present detailed results, including standard deviation, for the main experiments from the paper. These results are presented in Table~\ref{tab:imagenet_mean_std1} and Table~\ref{tab:imagenet_mean_std2}.

\begin{table*}[b]
\begin{subtable}{\textwidth}
\centering
\resizebox{\textwidth}{!}{%
    \begin{tabular}{c c|c c c c c c c c c c c}
    \hline
        \multirow{2}{*}{} & \multirow{2}{*}{Method} & \multicolumn{11}{|c}{acquisition round} \\
        & & \multicolumn{1}{|c}{0} & \multicolumn{1}{c}{1} & \multicolumn{1}{c}{2} & \multicolumn{1}{c}{3} & \multicolumn{1}{c}{4} & \multicolumn{1}{c}{5} 
        & \multicolumn{1}{c}{6} & \multicolumn{1}{c}{7} & \multicolumn{1}{c}{8} & \multicolumn{1}{c}{9} & \multicolumn{1}{c}{10}\\
    \hline
    \hline
    & Ours & 45.76 & 55.47 & 60.53 & 63.15 & 65.22 & 66.05 & 68.69 & 69.46 & 70.48 & 71.97 & 72.22 \\
    &              & ($\pm$3.27) & ($\pm$1.98) & ($\pm$2.04) & ($\pm$1.48) & ($\pm$0.64) & ($\pm$0.95) & ($\pm$0.90) & ($\pm$0.81) & ($\pm$1.32) & ($\pm$0.53) & ($\pm$1.11) \\
    & Ours w/o semi & 45.76 & 52.03 & 55.62 & 57.97 & 59.34 & 61.92 & 62.34 & 64.26 & 66.18 & 64.59 & 67.55 \\
    &              & ($\pm$3.27) & ($\pm$1.87) & ($\pm$1.10) & ($\pm$2.26) & ($\pm$1.25) & ($\pm$2.19) & ($\pm$1.41) & ($\pm$1.21) & ($\pm$0.83) & ($\pm$1.42) & ($\pm$1.16) \\
    & CCAL & 45.79 & 50.67 & 53.70 & 55.41 & 56.99 & 58.83 & 60.27 & 61.12 & 62.67 & 64.53 & 65.90 \\
    &      & ($\pm$2.72)	& ($\pm$1.11)	& ($\pm$2.00)	& ($\pm$2.17)	& ($\pm$1.94)	& ($\pm$1.19)	& ($\pm$1.40)	& ($\pm$1.53)	& ($\pm$1.64)	& ($\pm$0.26)	& ($\pm$0.71) \\
    & LfOSA & 44.96 & 48.29 & 51.84 & 54.02 & 55.28 & 55.92 & 58.11 & 59.52 & 60.03 & 62.37 & 63.06 \\
    &      & ($\pm$2.39) & ($\pm$2.12) & ($\pm$2.21) & ($\pm$0.74) & ($\pm$0.97) & ($\pm$2.04) & ($\pm$1.83) & ($\pm$1.40) & ($\pm$1.23) & ($\pm$1.60) & ($\pm$1.52) \\
    & MQNet & 44.96 & 50.54 & 54.19 & 55.97 & 59.41 & 60.67 & 61.87 & 62.72 & 64.05 & 65.89 & 65.38 \\
    &      & ($\pm$2.39) & ($\pm$1.60) & ($\pm$1.43) & ($\pm$2.07) & ($\pm$1.45) & ($\pm$1.80) & ($\pm$2.09) & ($\pm$1.78) & ($\pm$1.32) & ($\pm$1.33) & ($\pm$0.91) \\
    & SIMILAR & 45.42 & 51.17 & 55.74 & 57.23 & 60.64 & 61.06 & 62.58 & 62.53 & 64.61 & 65.95 & 65.87 \\
    &         & ($\pm$3.46) & ($\pm$2.99) & ($\pm$0.76) & ($\pm$1.76) & ($\pm$1.14) & ($\pm$0.75) & ($\pm$1.48) & ($\pm$1.57) & ($\pm$0.58) & ($\pm$1.36) & ($\pm$0.70) \\
    & Random & 44.96 & 52.53 & 55.63 & 58.83 & 61.89 & 61.41 & 62.64 & 63.31 & 63.82 & 65.41 & 65.52 \\
    &        & ($\pm$2.39) & ($\pm$0.92) & ($\pm$1.43) & ($\pm$1.35) & ($\pm$1.42) & ($\pm$1.66) & ($\pm$1.17) & ($\pm$0.94) & ($\pm$1.53) & ($\pm$1.96) & ($\pm$1.03) \\
    & BADGE & 44.96 & 52.05 & 56.13 & 58.51 & 60.19 & 61.55 & 63.17 & 64.18 & 64.99 & 67.02 & 68.59 \\
    &       & ($\pm$2.39) & ($\pm$2.64) & ($\pm$1.23) & ($\pm$1.92) & ($\pm$1.56) & ($\pm$1.16) & ($\pm$1.14) & ($\pm$1.71) & ($\pm$1.65) & ($\pm$0.78) & ($\pm$0.70) \\
    \hline
    \hline

\end{tabular}
}
\caption{Results for 0.0 outlier ratio.}\label{tab:imagenet_0}
\end{subtable}
\hfill
\begin{subtable}{\textwidth}
\centering
\resizebox{\textwidth}{!}{%
    \begin{tabular}{c c|c c c c c c c c c c c}
    \hline
        \multirow{2}{*}{} & \multirow{2}{*}{Method} & \multicolumn{11}{|c}{acquisition round} \\
        & & \multicolumn{1}{|c}{0} & \multicolumn{1}{c}{1} & \multicolumn{1}{c}{2} & \multicolumn{1}{c}{3} & \multicolumn{1}{c}{4} & \multicolumn{1}{c}{5} 
        & \multicolumn{1}{c}{6} & \multicolumn{1}{c}{7} & \multicolumn{1}{c}{8} & \multicolumn{1}{c}{9} & \multicolumn{1}{c}{10}\\
    \hline
    \hline
    & Ours & 41.12 & 54.85 & 60.03 & 62.38 & 65.71 & 66.91 & 67.42 & 69.87 & 72.02 & 72.10 & 73.22 \\
    &              & ($\pm$2.96) & ($\pm$2.61) & ($\pm$1.05) & ($\pm$0.84) & ($\pm$0.60) & ($\pm$0.66) & ($\pm$1.05) & ($\pm$0.74) & ($\pm$1.07) & ($\pm$1.26) & ($\pm$0.78) \\
    & Ours w/o semi & 41.12 & 47.41 & 54.29 & 56.38 & 58.37 & 59.44 & 62.86 & 61.17 & 63.74 & 64.35 & 64.66 \\
    &              & ($\pm$2.96) & ($\pm$2.90) & ($\pm$2.10) & ($\pm$1.53) & ($\pm$2.33) & ($\pm$1.87) & ($\pm$2.16) & ($\pm$2.91) & ($\pm$2.01) & ($\pm$2.06) & ($\pm$3.09) \\
    & CCAL & 41.71 & 45.89 & 49.82 & 54.51 & 55.41 & 57.49 & 59.28 & 61.41 & 63.12 & 62.43 & 64.51 \\
    &      & ($\pm$2.88)	& ($\pm$2.97)	& ($\pm$1.09)	& ($\pm$1.25)	& ($\pm$2.18)	& ($\pm$1.38)	& ($\pm$1.92)	& ($\pm$0.91)	& ($\pm$1.23)	& ($\pm$1.79)	& ($\pm$0.98) \\
    & LfOSA & 41.26 & 46.61 & 48.74 & 50.74 & 52.78 & 54.19 & 56.66 & 58.77 & 60.10 & 61.71 & 62.74 \\
    &      & ($\pm$2.13) & ($\pm$3.65) & ($\pm$2.94) & ($\pm$1.02) & ($\pm$1.79) & ($\pm$2.47) & ($\pm$2.84) & ($\pm$1.79) & ($\pm$1.60) & ($\pm$1.18) & ($\pm$1.51) \\
    & MQNet & 41.26 & 46.66 & 51.39 & 56.30 & 57.42 & 59.36 & 61.74 & 62.77 & 62.22 & 63.87 & 66.37 \\
    &      & ($\pm$2.13) & ($\pm$2.95) & ($\pm$3.71) & ($\pm$0.77) & ($\pm$2.36) & ($\pm$2.90) & ($\pm$0.63) & ($\pm$2.14) & ($\pm$2.24) & ($\pm$1.73) & ($\pm$1.77) \\
    & SIMILAR & 41.41 & 47.68 & 50.75 & 54.78 & 58.00 & 58.96 & 60.59 & 61.81 & 63.79 & 63.95 & 64.98 \\
    &         & ($\pm$2.87) & ($\pm$3.22) & ($\pm$3.20) & ($\pm$1.16) & ($\pm$0.95) & ($\pm$1.79) & ($\pm$1.06) & ($\pm$2.15) & ($\pm$1.29) & ($\pm$1.21) & ($\pm$1.31) \\
    & Random & 41.26 & 49.98 & 52.21 & 55.66 & 58.27 & 61.30 & 61.62 & 64.05 & 61.74 & 65.55 & 64.24 \\
    &        & ($\pm$2.13) & ($\pm$2.68) & ($\pm$1.67) & ($\pm$3.07) & ($\pm$3.66) & ($\pm$1.18) & ($\pm$2.26) & ($\pm$1.76) & ($\pm$2.33) & ($\pm$1.04) & ($\pm$1.99) \\
    & BADGE & 41.26 & 48.72 & 51.78 & 55.89 & 58.03 & 60.93 & 61.54 & 63.84 & 64.75 & 65.52 & 66.45 \\
    &       & ($\pm$2.13) & ($\pm$2.89) & ($\pm$1.38) & ($\pm$1.84) & ($\pm$1.96) & ($\pm$2.70) & ($\pm$2.40) & ($\pm$1.65) & ($\pm$0.85) & ($\pm$0.92) & ($\pm$2.01) \\
    \hline
    \hline

\end{tabular}
}
\caption{Results for 0.05 outlier ratio.}\label{tab:imagenet_005}
\end{subtable}
\hfill
\begin{subtable}{\textwidth}
\centering
\resizebox{\textwidth}{!}{%
    \begin{tabular}{c c|c c c c c c c c c c c}
    \hline
        \multirow{2}{*}{} & \multirow{2}{*}{Method} & \multicolumn{11}{|c}{acquisition round} \\
        & & \multicolumn{1}{|c}{0} & \multicolumn{1}{c}{1} & \multicolumn{1}{c}{2} & \multicolumn{1}{c}{3} & \multicolumn{1}{c}{4} & \multicolumn{1}{c}{5} 
        & \multicolumn{1}{c}{6} & \multicolumn{1}{c}{7} & \multicolumn{1}{c}{8} & \multicolumn{1}{c}{9} & \multicolumn{1}{c}{10}\\
    \hline
    \hline
    & Ours & 40.66 & 52.66 & 57.47 & 61.65 & 63.90 & 66.13 & 67.28 & 69.57 & 70.64 & 71.73 & 71.36 \\
    &              & ($\pm$2.89) & ($\pm$1.79) & ($\pm$2.00) & ($\pm$0.74) & ($\pm$1.54) & ($\pm$1.55) & ($\pm$0.86) & ($\pm$1.39) & ($\pm$0.90) & ($\pm$1.37) & ($\pm$0.79) \\
    & Ours w/o semi & 40.66 & 46.48 & 52.38 & 53.38 & 57.50 & 58.85 & 61.12 & 63.01 & 63.52 & 63.74 & 65.20 \\
    &              & ($\pm$2.89) & ($\pm$2.23) & ($\pm$2.21) & ($\pm$2.30) & ($\pm$1.82) & ($\pm$2.16) & ($\pm$1.22) & ($\pm$0.93) & ($\pm$1.88) & ($\pm$1.15) & ($\pm$1.59) \\
    & CCAL & 40.05 & 46.61 & 52.88 & 54.08 & 56.59 & 56.90 & 59.07 & 60.29 & 61.15 & 63.26 & 61.84 \\
    &      & ($\pm$2.58)	& ($\pm$1.45)	& ($\pm$1.22)	& ($\pm$0.49)	& ($\pm$1.47)	& ($\pm$2.42)	& ($\pm$0.96)	& ($\pm$2.14)	& ($\pm$1.91)	& ($\pm$1.43)	& ($\pm$1.86) \\
    & LfOSA & 40.94 & 43.74 & 48.35 & 50.13 & 50.83 & 53.71 & 56.94 & 57.20 & 59.22 & 61.01 & 62.62 \\
    &      & ($\pm$2.90) & ($\pm$3.19) & ($\pm$2.02) & ($\pm$1.43) & ($\pm$1.88) & ($\pm$1.92) & ($\pm$0.83) & ($\pm$2.63) & ($\pm$1.63) & ($\pm$2.08) & ($\pm$1.07) \\
    & MQNet & 40.94 & 45.94 & 51.14 & 53.09 & 54.40 & 57.15 & 56.59 & 57.74 & 60.11 & 61.79 & 62.37 \\
    &      & ($\pm$2.90) & ($\pm$1.64) & ($\pm$0.48) & ($\pm$2.32) & ($\pm$1.80) & ($\pm$2.29) & ($\pm$1.57) & ($\pm$1.91) & ($\pm$3.91) & ($\pm$1.85) & ($\pm$2.03) \\
    & SIMILAR & 40.91 & 47.38 & 51.95 & 53.90 & 55.34 & 57.97 & 61.41 & 62.16 & 60.96 & 61.87 & 63.58 \\
    &         & ($\pm$2.95) & ($\pm$1.88) & ($\pm$2.91) & ($\pm$2.23) & ($\pm$0.74) & ($\pm$1.51) & ($\pm$1.30) & ($\pm$1.40) & ($\pm$2.87) & ($\pm$1.72) & ($\pm$2.44) \\
    & Random & 40.94 & 46.24 & 52.06 & 52.38 & 56.51 & 58.26 & 57.90 & 58.90 & 63.28 & 62.18 & 64.26 \\
    &        & ($\pm$2.90) & ($\pm$0.85) & ($\pm$2.42) & ($\pm$1.95) & ($\pm$3.02) & ($\pm$1.47) & ($\pm$2.63) & ($\pm$1.81) & ($\pm$0.62) & ($\pm$1.89) & ($\pm$1.47) \\
    & BADGE & 40.94 & 46.69 & 51.15 & 52.99 & 56.66 & 58.78 & 58.96 & 61.02 & 61.81 & 63.36 & 63.81 \\
    &       & ($\pm$2.90) & ($\pm$2.63) & ($\pm$1.70) & ($\pm$1.99) & ($\pm$0.96) & ($\pm$1.48) & ($\pm$2.40) & ($\pm$3.09) & ($\pm$2.53) & ($\pm$2.31) & ($\pm$1.85) \\
    \hline
    \hline

\end{tabular}
}
\caption{Results for 0.2 outlier ratio.}\label{tab:imagenet_02}
\end{subtable}
\caption{Mean and standard deviation for different methods on ImageNet dataset. \label{tab:imagenet_mean_std1} }
\end{table*}
\begin{table*}[b]
\begin{subtable}{\textwidth}
\centering
\resizebox{\textwidth}{!}{%
    \begin{tabular}{c c|c c c c c c c c c c c}
    \hline
        \multirow{2}{*}{} & \multirow{2}{*}{Method} & \multicolumn{11}{|c}{acquisition round} \\
        & & \multicolumn{1}{|c}{0} & \multicolumn{1}{c}{1} & \multicolumn{1}{c}{2} & \multicolumn{1}{c}{3} & \multicolumn{1}{c}{4} & \multicolumn{1}{c}{5} 
        & \multicolumn{1}{c}{6} & \multicolumn{1}{c}{7} & \multicolumn{1}{c}{8} & \multicolumn{1}{c}{9} & \multicolumn{1}{c}{10}\\
    \hline
    \hline
    & Ours & 41.98 & 52.66 & 56.82 & 61.44 & 64.30 & 65.94 & 68.88 & 69.55 & 71.07 & 71.10 & 73.06 \\
    &              & ($\pm$3.60) & ($\pm$1.68) & ($\pm$0.97) & ($\pm$2.37) & ($\pm$1.47) & ($\pm$1.27) & ($\pm$0.73) & ($\pm$1.34) & ($\pm$0.49) & ($\pm$0.63) & ($\pm$1.28) \\
    & Ours w/o semi & 41.98 & 47.42 & 53.12 & 56.26 & 58.51 & 60.90 & 61.30 & 63.25 & 63.86 & 64.37 & 66.16 \\
    &              & ($\pm$3.60) & ($\pm$2.83) & ($\pm$2.07) & ($\pm$1.00) & ($\pm$3.63) & ($\pm$2.09) & ($\pm$1.13) & ($\pm$1.87) & ($\pm$3.89) & ($\pm$0.86) & ($\pm$2.33) \\
    & CCAL & 43.79 & 47.76 & 50.45 & 51.70 & 54.14 & 56.80 & 59.79 & 59.55 & 61.09 & 60.22 & 63.12 \\
    &      & ($\pm$1.77)	& ($\pm$2.08)	& ($\pm$2.70)	& ($\pm$1.84)	& ($\pm$1.95)	& ($\pm$2.44)	& ($\pm$0.95)	& ($\pm$2.17)	& ($\pm$1.83)	& ($\pm$2.18)	& ($\pm$1.09) \\
    & LfOSA & 41.76 & 45.87 & 47.97 & 51.94 & 51.87 & 55.36 & 56.02 & 56.05 & 59.42 & 60.38 & 61.94 \\
    &      & ($\pm$2.62) & ($\pm$3.44) & ($\pm$1.92) & ($\pm$1.31) & ($\pm$1.37) & ($\pm$2.60) & ($\pm$2.32) & ($\pm$2.85) & ($\pm$1.61) & ($\pm$0.97) & ($\pm$1.52) \\
    & MQNet & 41.76 & 47.87 & 50.13 & 53.87 & 54.67 & 55.42 & 56.93 & 60.37 & 59.42 & 62.46 & 62.10 \\
    &      & ($\pm$2.62) & ($\pm$1.45) & ($\pm$3.47) & ($\pm$3.13) & ($\pm$2.08) & ($\pm$1.86) & ($\pm$2.67) & ($\pm$1.14) & ($\pm$1.56) & ($\pm$0.69) & ($\pm$2.18) \\
    & SIMILAR & 41.84 & 47.65 & 52.91 & 54.93 & 57.20 & 58.56 & 59.09 & 61.47 & 63.33 & 63.90 & 65.14 \\
    &         & ($\pm$3.62) & ($\pm$0.99) & ($\pm$1.66) & ($\pm$2.26) & ($\pm$1.11) & ($\pm$2.22) & ($\pm$1.37) & ($\pm$1.27) & ($\pm$1.63) & ($\pm$0.55) & ($\pm$1.21) \\
    & Random & 41.76 & 46.91 & 49.23 & 52.26 & 55.17 & 56.93 & 58.05 & 58.35 & 60.42 & 61.28 & 61.33 \\
    &        & ($\pm$2.62) & ($\pm$2.36) & ($\pm$1.66) & ($\pm$1.94) & ($\pm$1.30) & ($\pm$1.73) & ($\pm$1.80) & ($\pm$1.37) & ($\pm$0.69) & ($\pm$1.81) & ($\pm$2.28) \\
    & BADGE & 41.76 & 47.60 & 49.76 & 51.55 & 54.50 & 56.45 & 57.82 & 56.80 & 59.94 & 61.78 & 61.89 \\
    &       & ($\pm$2.62) & ($\pm$1.56) & ($\pm$2.04) & ($\pm$4.35) & ($\pm$3.41) & ($\pm$2.25) & ($\pm$1.23) & ($\pm$3.02) & ($\pm$1.32) & ($\pm$2.09) & ($\pm$1.54) \\
    \hline
    \hline

\end{tabular}
}
\caption{Results for 0.5 outlier ratio.}\label{tab:imagenet_05}
\end{subtable}
\hfill
\begin{subtable}{\textwidth}
\centering
\resizebox{\textwidth}{!}{%
    \centering
    \begin{tabular}{c c|c c c c c c c c c c c}
    \hline
        \multirow{2}{*}{} & \multirow{2}{*}{Method} & \multicolumn{11}{|c}{acquisition round} \\
        & & \multicolumn{1}{|c}{0} & \multicolumn{1}{c}{1} & \multicolumn{1}{c}{2} & \multicolumn{1}{c}{3} & \multicolumn{1}{c}{4} & \multicolumn{1}{c}{5} 
        & \multicolumn{1}{c}{6} & \multicolumn{1}{c}{7} & \multicolumn{1}{c}{8} & \multicolumn{1}{c}{9} & \multicolumn{1}{c}{10}\\
    \hline
    \hline
    & Ours & 41.12 & 48.48 & 52.77 & 56.69 & 61.15 & 62.96 & 63.57 & 66.43 & 67.3 & 67.68 & 68.98 \\
    &              & ($\pm$ 2.96) & ($\pm$1.02) & ($\pm$1.86) & ($\pm$0.51) & ($\pm$1.57) & ($\pm$1.15) & ($\pm$1.08) & ($\pm$1.27) & ($\pm$1.92) & ($\pm$1.81) & ($\pm$1.28) \\
    & Ours w/o semi & 41.12 & 43.1 & 48.7 & 49.92 & 54.98 & 56.58 & 56.83 & 59.95 & 61.58 & 61.57 & 63.02 \\
    &              & ($\pm$ 2.96) & ($\pm$1.56) & ($\pm$1.35) & ($\pm$2.30) & ($\pm$1.81) & ($\pm$3.48) & ($\pm$2.70) & ($\pm$0.89) & ($\pm$1.32) & ($\pm$2.38) & ($\pm$1.30) \\
    & CCAL & 41.71 & 45.57 & 47.01 & 49.10 & 48.66 & 50.90 & 51.47 & 53.38 & 55.07 & 54.66 & 53.26 \\
    &      & ($\pm$2.88) & ($\pm$1.60) & ($\pm$1.62) & ($\pm$1.50) & ($\pm$1.78) & ($\pm$1.42) & ($\pm$1.57) & ($\pm$1.05) & ($\pm$2.63) & ($\pm$1.17) & ($\pm$2.71) \\
    & LfOSA & 41.26 & 44.13 & 47.39 & 48.75 & 51.26 & 52.22 & 54.51 & 54.59 & 56.94 & 58.03 & 60.05 \\
    &      & ($\pm$2.13) & ($\pm$2.63) & ($\pm$2.78) & ($\pm$1.86) & ($\pm$1.10) & ($\pm$1.56) & ($\pm$1.32) & ($\pm$2.02) & ($\pm$2.60) & ($\pm$3.09) & ($\pm$1.66) \\
    & MQNet & 41.26 & 44.19 & 46.34 & 50.08 & 50.24 & 49.79 & 50.93 & 52.29 & 53.09 & 53.65 & 54.46 \\
    &      & ($\pm$2.13) & ($\pm$1.40) & ($\pm$1.28) & ($\pm$2.27) & ($\pm$1.74) & ($\pm$1.78) & ($\pm$2.12) & ($\pm$1.79) & ($\pm$1.47) & ($\pm$1.93) & ($\pm$1.43) \\
    & SIMILAR & 41.41 & 46.69 & 48.85 & 50.11 & 54.13 & 55.22 & 58.94 & 55.92 & 59.34 & 59.47 & 61.84 \\
    &         & ($\pm$2.87) & ($\pm$2.25) & ($\pm$2.68) & ($\pm$1.30) & ($\pm$1.38) & ($\pm$0.94) & ($\pm$0.94) & ($\pm$3.43) & ($\pm$1.65) & ($\pm$1.68) & ($\pm$1.59) \\
    & Random & 41.26 & 43.50 & 45.06 & 46.34 & 47.50 & 48.74 & 50.93 & 51.62 & 52.26 & 52.99 & 53.30 \\
    &        & ($\pm$2.13) & ($\pm$3.14) & ($\pm$2.04) & ($\pm$1.33) & ($\pm$2.06) & ($\pm$2.30) & ($\pm$2.79) & ($\pm$1.44) & ($\pm$1.04) & ($\pm$4.22) & ($\pm$1.52) \\
    & BADGE & 41.26 & 44.72 & 45.89 & 47.10 & 46.54 & 49.15 & 51.65 & 49.49 & 52.66 & 53.14 & 55.09 \\
    &       & ($\pm$2.13) & ($\pm$1.31) & ($\pm$1.82) & ($\pm$1.85) & ($\pm$2.55) & ($\pm$1.62) & ($\pm$0.66) & ($\pm$3.42) & ($\pm$2.20) & ($\pm$1.93) & ($\pm$1.59) \\
    \hline
    \hline

\end{tabular}
}

\caption{Results for 0.8 outlier ratio.}\label{tab:imagenet_08}
\end{subtable}
\hfill
\begin{subtable}{\textwidth}
\centering
\resizebox{\textwidth}{!}{%
    \begin{tabular}{c c|c c c c c c c c c c c}
    \hline
        \multirow{2}{*}{} & \multirow{2}{*}{Method} & \multicolumn{11}{|c}{acquisition round} \\
        & & \multicolumn{1}{|c}{0} & \multicolumn{1}{c}{1} & \multicolumn{1}{c}{2} & \multicolumn{1}{c}{3} & \multicolumn{1}{c}{4} & \multicolumn{1}{c}{5} 
        & \multicolumn{1}{c}{6} & \multicolumn{1}{c}{7} & \multicolumn{1}{c}{8} & \multicolumn{1}{c}{9} & \multicolumn{1}{c}{10}\\
    \hline
    \hline
    & Ours & 41.12 & 48.00 & 50.00 & 53.15 & 56.21 & 58.35 & 61.30 & 62.00 & 64.80 & 64.78 & 65.22 \\
    &              & ($\pm$2.96) & ($\pm$0.85) & ($\pm$2.46) & ($\pm$1.79) & ($\pm$1.09) & ($\pm$2.26) & ($\pm$1.10) & ($\pm$0.25) & ($\pm$0.79) & ($\pm$0.72) & ($\pm$1.06) \\
    & Ours w/o semi & 41.12 & 42.27 & 44.90 & 49.94 & 51.82 & 54.66 & 54.34 & 57.41 & 58.86 & 59.26 & 62.14 \\
    &              & ($\pm$2.96) & ($\pm$2.12) & ($\pm$1.84) & ($\pm$0.94) & ($\pm$2.08) & ($\pm$2.69) & ($\pm$2.88) & ($\pm$1.25) & ($\pm$2.09) & ($\pm$2.61) & ($\pm$0.84) \\
    & CCAL & 41.71 & 43.38 & 45.23 & 45.82 & 47.52 & 47.66 & 47.65 & 47.78 & 49.23 & 51.38 & 50.45 \\
    &      & ($\pm$2.88)	& ($\pm$1.99)	& ($\pm$1.32)	& ($\pm$2.30)	& ($\pm$2.40)	& ($\pm$2.00)	& ($\pm$1.96)	& ($\pm$1.38)	& ($\pm$2.52)	& ($\pm$1.41)	& ($\pm$2.37) \\
    & LfOSA & 41.26 & 44.19 & 45.20 & 48.88 & 49.73 & 51.12 & 51.90 & 52.72 & 55.52 & 56.16 & 57.90 \\
    &      & ($\pm$2.13) & ($\pm$2.18) & ($\pm$1.59) & ($\pm$1.51) & ($\pm$1.67) & ($\pm$1.28) & ($\pm$1.66) & ($\pm$1.77) & ($\pm$1.21) & ($\pm$1.10) & ($\pm$1.24) \\
    & MQNet & 41.26 & 42.03 & 45.81 & 45.50 & 47.10 & 48.83 & 47.14 & 48.72 & 47.95 & 50.29 & 51.17 \\
    &      & ($\pm$2.13) & ($\pm$2.56) & ($\pm$2.56) & ($\pm$2.28) & ($\pm$1.52) & ($\pm$1.40) & ($\pm$1.13) & ($\pm$1.34) & ($\pm$4.34) & ($\pm$1.79) & ($\pm$2.36) \\
    & Random & 41.26 & 42.51 & 43.30 & 43.60 & 45.86 & 45.70 & 46.83 & 47.62 & 48.88 & 49.81 & 51.70 \\
    &        & ($\pm$2.13) & ($\pm$1.89) & ($\pm$3.84) & ($\pm$2.21) & ($\pm$1.87) & ($\pm$2.38) & ($\pm$1.55) & ($\pm$0.98) & ($\pm$3.09) & ($\pm$1.87) & ($\pm$1.92) \\
    & BADGE & 41.26 & 43.68 & 43.94 & 45.44 & 45.54 & 46.85 & 47.92 & 47.60 & 47.18 & 46.86 & 49.07 \\
    &       & ($\pm$2.13) & ($\pm$2.86) & ($\pm$2.45) & ($\pm$0.96) & ($\pm$0.45) & ($\pm$2.82) & ($\pm$1.32) & ($\pm$0.98) & ($\pm$2.56) & ($\pm$3.91) & ($\pm$1.59) \\
    \hline
    \hline

\end{tabular}
}
\caption{Results for 0.9 outlier ratio.}\label{tab:imagenet_09}
\end{subtable}
\caption{Mean and standard deviation for different methods on ImageNet dataset. \label{tab:imagenet_mean_std2} }
\end{table*}

\section{Implementation details}

For the backbone of all experiments, we use ResNet18~\cite{hzr+16}. For CIFAR100 and TinyImageNet experiments, we use the variant commonly used for CIFAR experiments. It is standard practice to use this variant~\cite{tavaal,similar} which uses a kernel of size 3 and stride 1 instead of 7 and 2, respectively, in the first convolutional layer\footnote{This architecture is used for SLL by CCAL, but not for the classifier, even though we found it to be beneficial.}. For ImageNet experiments, we use the standard version with a kernel size of 7 and stride 2 in the first convolutional layer. SSL pre-training is performed for 700 epochs using a batch of size 32, 64, and 100 for CIFAR100, TinyImageNet, and ImageNet, respectively, initial learning rate equal to 1e-1 with cosine annealing and SGD optimizer. The result is used as initialization for classifier training, which is performed for 10 epochs using a batch of size 32, learning rate equal to 5e-4, and Adam optimizer for the training on the labeled set. In the experiments, this setup is fixed for all methods we compare with. For the semi-supervised training, we continue training from the point where training on the labeled set stopped. We do this for 3 epochs, where we consider one full pass through the unlabeled set as the epoch. We use a batch size of 512, where half of the batch comes from the unlabeled set and the other half comes from the labeled set. During training, we use random horizontal flipping as the augmentation on CIFAR100 and TinyImageNet, while on ImageNet, we first perform random resized cropping and then random horizontal flipping. Pseudo-code of our method is presented in Algorithm~\ref{alg:main}.

We run CoreSet, BADGE, CCAL, SIMILAR, and MQNet using the provided implementations \footnote{\url{https://github.com/RUC-DWBI-ML/CCAL}\\~~\url{https://github.com/decile-team/distil}\\~~\url{https://github.com/kaist-dmlab/MQNet}}, after integrating them into our implementation framework. We implement LfOSA by ourselves.
\section{Benchmark details}
\looseness=-1
The original CIFAR100 consists of 100 categories. We use 20 of them as inlier classes, and the rest are used to form the outlier class. The former correspond to large omnivores and herbivores, medium-sized mammals, and small mammals.
This particular way of splitting classes is performed in prior work, but without publicly sharing the list of images per split~\cite{ccal}. Therefore, we adopt the same class splits and define our own image splits, which we will publicly share. 
The test set is formed by examples coming from the test split and contains only images from inlier classes giving us 2000 images.

\begin{algorithm}[H]
    \centering
    \footnotesize
\algnewcommand\False{\textbf{false}\space}
\algnewcommand\True{\textbf{true}\space}
\algnewcommand\Andc{\textbf{and}\space}
\algrenewcommand\algorithmicindent{0.9em}%

\begin{spacing}{1.0}
    \begin{algorithmic}[1]
        \Procedure{AL}{labeled set $L_0$, unlabeled set $U_0$, do-semi, do-filtering}
            \State $f_{\text{init}} \gets$ SSL on $L_0 \cup U_0$ \comment{self-supervised pre-training}
            \For{$t \in [0, \ldots, T]$} \comment{active learning rounds}
                \For{$i \in [1, \ldots, M]$} \comment{supervised training, $M$ models}
                    \State $f_{t_{i}} \gets \arg\min_{f} \cL(L_t; f)$ \comment{start from $f_{\text{init}}$, train $\cL$}
                \EndFor
                \If {do-semi is \True \Andc $t \neq 0$} \comment{semi-supervision}
                    \State \algorithmicfor {~$x \in U_{0}$} \algorithmicdo  {~$\hat{y}_t(x) \gets \arg\max_{j} F_{t}(x)_{j}$}\ \comment{pseudo-label}
                    \State \algorithmicfor {~$x \in U_{0}$} \algorithmicdo  {~$w_t(x) \gets 1 - \frac{H\left(F_{t}(x)\right)}{\log (K+1)}$}\ \comment{weights}
                    \For{$i \in [1, \ldots, M]$} \comment{semi-supervised training, $M$ models}
                        \State $f^{'}_{t_{i}} \gets \arg\min_{f} \cL_{\text{semi}}(L_t, U_t; f)$ \comment{train longer with $\cL_\text{semi}$}
                    \EndFor
                \EndIf
                \For {$x \in U_t$} \comment{loop to estimate acquisition score}
                    \If {$t = 0$}
                        \State $a_{t}(x) \sim \mathcal{U}_{[0,1]}$ \comment{random chance}
                    \Else
                        \State $\tilde{a}_{t}(x) \gets 1-\frac{\left|\left\{i:~\hat{y}^{\prime}_{t_i}(x) =  \hat{y}^{\prime}_t(x) \right\}\right|}{M}$ \comment{VR score}
                        \If {do-filtering is \True}
                            \State $a_{t}(x) \gets \tilde{a}_{t}(x)\mathds{1}_{\hat{y}_t(x)\neq C_o}$ \comment{filtering}
                        \Else
                            \State $a_{t}(x) \gets \tilde{a}_{t}(x)$ \comment{no filtering}
                        \EndIf
                    \EndIf
                \EndFor
                \State $A_t \gets \topB \{a_{t}(x): x\in U_{t}\}$ \comment{example selection based on largest score}
                \State $\text{annotate}(A_t)$ \comment{annotators assign labels}
                \State $L_{t+1} \gets L_{t} \cup A_t$ \comment{update the labeled set}
                \State $U_{t+1} \gets U_{t} \setminus A_t$ \comment{update the unlabeled set}
            \EndFor
        \EndProcedure
    \end{algorithmic}
\end{spacing}
    \caption{Overview of the approach.\label{alg:main}}
\end{algorithm}

\looseness=-1
The original TinyImageNet consists of 200 categories. We use 25 categories corresponding to land animals as inlier classes, and the rest are used to form the outlier class. The test set is formed by examples from the validation split and contains only images from inlier classes, giving us 1250 test images.

\vspace{5pt}

We provide the inlier/outlier class splits for CIFAR100, TinyImageNet, and ImageNet datasets. While CIFAR100 splits are obtained from CCAL~\cite{ccal}, TinyImageNet, and ImageNet splits are created from scratch for our work.
The ids for classes used as inliers are listed below, while the ids of outlier classes will be released with the code.

\begin{enumerate}
    \setlength\itemsep{-8pt}
    \item CIFAR100: 3, 42, 43, 88, 97, 15, 19, 21, 32, 39, 35, 63, 64, 66, 75, 37, 50, 65, 74, 80
    \item TinyImageNet: 29, 54, 114, 159, 171, 197, 94, 174, 192, 28, 1, 11, 5, 24, 83, 128, 82, 108, 118, 98, 180, 62, 163, 111, 78
    \item ImageNet: \\ n02085620, n02086240, n02086910, n02087046, n02089867, n02089973, n02090622, n02091831, n02093428, n02099849, n02100583, n02104029, n02105505, n02106550, n02107142, n02108089, n02109047, n02113799, n02113978, n02114855, n02116738, n02119022, n02123045, n02138441, n02326432
\end{enumerate}

\end{document}